\documentclass[journal]{IEEEtran}
\usepackage{graphicx}
\usepackage{amsfonts}

\usepackage[english]{babel}
\usepackage{amsmath, amssymb}
\usepackage{float}
\usepackage{makecell}
\usepackage{multirow}
\usepackage{cite}
\usepackage{amsmath,amssymb,amsfonts}
\usepackage{algorithmic}
\usepackage{graphicx}
\usepackage{textcomp}
\usepackage{xcolor}

\usepackage{subfigure}
\usepackage{epstopdf}

\usepackage{graphicx}
\usepackage{comment}
\usepackage{amsmath,amssymb} 
\usepackage{color}
\usepackage{enumerate}

\usepackage{booktabs}

\usepackage{soul}
\soulregister\cite7 
\soulregister\citep7 
\soulregister\citet7 
\soulregister\ref7 
\soulregister\pageref7 

\ifCLASSINFOpdf
\else
\fi
\begin{document}
%
\title{3D Hierarchical Refinement and Augmentation for Unsupervised Learning of Depth and Pose from Monocular Video}
%
%
%

\author{Guangming~Wang,
       Jiquan Zhong, Shijie Zhao, Wenhua Wu, Zhe Liu, and Hesheng Wang
        
\thanks{*
This work was supported in part by the Natural Science Foundation of China under Grant 62073222, U21A20480 and U1913204, in part by by the Science and Technology Commission of Shanghai Municipality under Grant 21511101900, in part by the Open Research Projects of Zhejiang Lab under Grant 2022NB0AB01, in part by grants from NVIDIA Corporation. Corresponding Author: Hesheng Wang.}
\thanks{G. Wang, J. Zhong, W. Wu, and H. Wang are with Department of Automation, Institute of Medical Robotics, Key Laboratory of System Control and Information Processing of Ministry of Education, Key Laboratory of Marine Intelligent Equipment and System of Ministry of Education, Shanghai Engineering Research Center of Intelligent Control and Management, Shanghai Jiao Tong University, Shanghai 200240, China.}
\thanks{S. Zhao is with the Department of Engineering Mechanics, Shanghai Jiao Tong University, Shanghai 200240, China.}%
\thanks{Z. Liu is with the Department of Computer Science and Technology, University of Cambridge, Cambridge, CB2 1TN, U.K.}
	
}

%
%

\markboth{Journal of \LaTeX\ Class Files,~Vol.~14, No.~8, August~2015}%
{Shell \MakeLowercase{\textit{et al.}}: Bare Demo of IEEEtran.cls for IEEE Journals}
%



\maketitle

\begin{abstract}
Depth and ego-motion estimations are essential for the localization and navigation of autonomous robots and autonomous driving. Recent studies make it possible to learn the per-pixel depth and ego-motion from the unlabeled monocular video.  A novel unsupervised training framework is proposed with 3D hierarchical refinement and augmentation using explicit 3D geometry. In this framework, the depth and pose estimations are hierarchically and mutually coupled to refine the estimated pose layer by layer. The intermediate view image is proposed and synthesized by warping the pixels in an image with the estimated depth and coarse pose. Then, the residual pose transformation can be estimated from the new view image and the image of the adjacent frame to refine the coarse pose. The iterative refinement is implemented in a differentiable manner in this paper, making the whole framework optimized uniformly.
Meanwhile, a new image augmentation method is proposed for the pose estimation by synthesizing a new view image, which creatively augments the pose in 3D space but gets a new augmented 2D image.
The experiments on KITTI demonstrate that our depth estimation achieves state-of-the-art performance and even surpasses recent approaches that utilize other auxiliary tasks. Our visual odometry outperforms all recent unsupervised monocular learning-based methods and achieves competitive performance to the geometry-based method, ORB-SLAM2 with back-end optimization.
\end{abstract}

\begin{IEEEkeywords}Monocular depth estimation, visual odometry, unsupervised learning, pose refinement, 3D augmentation.
\end{IEEEkeywords}

\IEEEpeerreviewmaketitle

\section{Introduction}\label{intro}

The depth and ego-motion (which also can be noted as pose) estimations ~\cite{chen2021fixing,tian2021depth,wang2021pwclo} are the basis of autonomous robot localization and navigation. The traditional monocular Simultaneous Localization and Mapping (SLAM) system needs initialization. The manual feature extraction is difficult to generalize in various environments, such as the environment with less texture.
Dense depth estimation~\cite{eigen2014depth,song2021monocular} based on deep learning has been developed with superior performance from only monocular camera. However, the ground truth of dense depth is expensive. Recently, unsupervised learning enables the depth and pose jointly learned without labels \cite{zhou2017unsupervised, GANVO, 8793622, bian2019depth, godard2019digging,chen2021fixing,tian2021depth}. After network training, depth network and pose network can also be evaluated separately so that depth estimation no longer relies on the existing pose estimation, and 3D information can be obtained only from a monocular image.
Moreover, image reconstruction is a key step in unsupervised joint learning of depth and pose. However, due to the occlusion, dynamics, and illumination variation, the image reconstruction is not always perfect. Many studies focus on the loss calculation and make the reconstruction error influenced less by the occlusion or dynamic objects in the process of network training \cite{8793622, bian2019depth, Gordon_2019_ICCV,godard2019digging}. Some works exploit joint learning with other auxiliary tasks to handle this challenge \cite{yang2018unsupervised, Yin_2018_CVPR, Zou_2018_ECCV, Ranjan_2019_CVPR,wang2020unsupervised,9156629}.

\begin{figure}[t]
	\centering
	\resizebox{0.95\columnwidth}{!}
	{
		\includegraphics[scale=1.00]{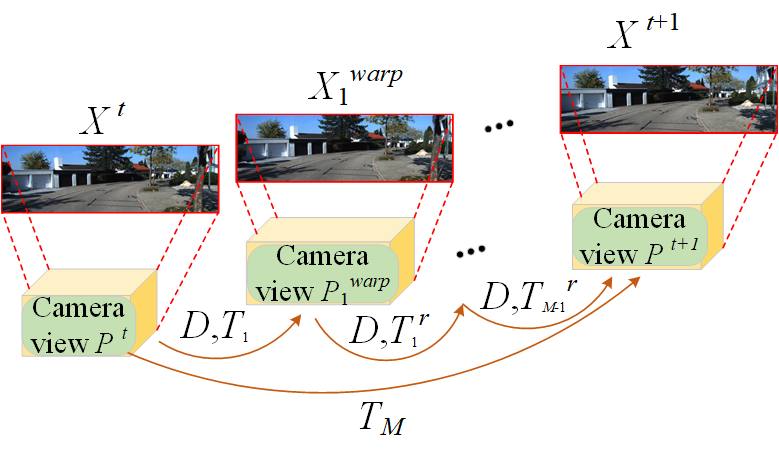}}
	\vspace{-4mm}
	\caption{
	The figure shows the process of pose estimation refinement. For details, given two adjacent frames $X^t$ and $X^{t+1}$, with their corresponding camera view $P^t$ and $P^{t+1}$, we utilize our 3D  hierarchical  refinement method introduced in Sec. \ref{multi} to synthesize transitional camera view $P^{warp}_m (m=1,2...M-1)$ and get residual pose $T^r_m (m=1,2...M-1)$. By multiplying $T_1,T^r_1...T^r_{M-1}$ step by step, we could refine our pose continuously. }
	\label{origin}
\end{figure}

In this paper,  we propose a new perspective to improve the performance of joint learning for depth and pose estimation. Specifically, the pose estimation
tends to degrade due to the considerable distance between two adjacent frames and the influence of the occlusion area. Therefore, we generate new intermediate views to 
enhance the accuracy of pose estimation, as shown in Fig. \ref{origin}. Due to the joint training, the depth network is also affected and improved simultaneously.
Meanwhile, the pose estimated by the coarse pose network can transform the predicted 3D point cloud in one frame to an adjacent frame based on the depth estimation results. Then, we generate masks from the operation above to solve the problem of occlusion and dynamic objects.  After the coarse pose transformation, the point cloud is reprojected onto the image plane of the adjacent frame, which makes it easier to match the pixels of adjacent frames. It is worth mentioning that the depth network has multi-scale depth outputs, and the method proposed in this paper also has multiple pose estimations. Therefore, there are many combinations to associate them and construct joint constraints. The best way is demonstrated through experiments and corresponding analysis in this paper.

In addition, the commonly used KITTI dataset \cite{doi:10.1177/0278364913491297} has only sample camera motion patterns \cite{wang2020tartanair}, which makes the trained pose network less generalized. Wang et al. \cite{wang2020tartanair} constructed a new dataset with complex and diverse motion patterns and changeable scenes. However, the dataset is virtual, and the domain gap is unavoidable \cite{tartanvo2020corl}. It is costly to simultaneously meet the complex and diverse motion patterns of the environment
in practice. To improve the generalization capability of the pose network, diverse motion patterns are required during training. Therefore, a new 6 Degrees-of-Freedom (DoF) pose augmentation method is proposed based on the depth estimation in the joint unsupervised training to synthesize new views from raw images, thereby enriching the training data and improving the performance of the pose network.

In general, the contributions of this paper are as follows:
\begin{itemize}
	\item Relying on the explicit 3D information from depth estimation, the concept of intermediate view synthesis is proposed in the joint unsupervised learning of depth and pose. A new joint unsupervised learning framework of depth and pose with the iterative view  synthesis and pose refinement is built to refine the estimated pose in an end-to-end fashion. 
	 
	\item  In the proposed framework, 6-DoF pose argumentation is proposed to synthesize a new view image, making the dataset for the training of the pose network cheaply expanded. To our best knowledge, this paper is the first to do the 6-DoF pose augmentation in visual odometry learning. The whole framework is implemented in a differentiable manner to optimize the depth and pose jointly.
	\item The mutual coupling learning of multi-scale depth estimation and multiple pose estimation is carefully considered and demonstrated by experiments on the KITTI dataset, which prove the proposed method achieves state-of-the-art performance on both depth and pose estimations. Besides, without relying on other auxiliary tasks and postprocessing, Our visual odometry achieves competitive performance to the geometry-based method, ORB-SLAM2~\cite{mur2017orb} with back-end optimization and closed-loop optimization.
\end{itemize}

The rest of this article is as follows. Section II summarizes some works related to this paper. Section III describes our method in detail. Section IV gives the details and results of the experiment. Finally, we conclude in section V.

\section{Related Work}
Deep learning has made remarkable achievements in depth and pose estimation, and we review the previous work in this section.

\subsection{Deep Supervised Learning}

Learning scene depth from a single image by Convolutional Neural Network (CNN) can date back to Eigen et al. \cite{eigen2014depth}.
They use two deep neural networks for depth estimation of images, one for global estimation and the other for refinement of local estimation. The output of the global network is part of the input of the local network to enhance depth refinement. In addition, they consider the relationship of the depth between adjacent pixels to reduce the prediction error.
Liu et al. \cite{liu2015learning}  express depth estimation as a continuous conditional random field (CRF) problem and propose a deep convolution neural field model, which can calculate gradient accurately because of the continuity of depth value.
Kuznietsov et al. \cite{Kuznietsov_2017_CVPR} propose a monocular depth map prediction method based on semi-supervised learning using sparse ground truth depth.
Guo et al. \cite{Guo_2018_ECCV} solve the gap between synthetic data and real data by using stereo matching as a proxy task.

\subsection{Self-supervised from Stereo Video}
Garg et al. \cite{garg2016unsupervised} propose the first end-to-end unsupervised convolutional neural network for depth estimation, which eliminates the need for large amounts of labeled data and greatly reduces the cost of training.
Image reconstruction alone leads to low-quality depth maps. Godard et al. \cite{godard2017unsupervised} propose a new kind of training loss that enhances consistency between the left and right images.
Zhan et al. \cite{zhan2018unsupervised} consider the information of consecutive frames and propose an additional feature reconstruction loss, which significantly improves the accuracy of depth estimation.
SGANVO \cite{8747446} proposes a novel pose network that uses the stacked generative adversarial structure and gets good performance due to its recurrent module across the network. 
Wong et al. \cite{wong2019bilateral} propose an adaptive weight to adjust the proportion of regularization loss and a bilateral cyclic consistency loss to solve the occlusion problem. They also built a two-branch decoder, with one branch predicting the initial disparity and the other refining it.

\begin{figure*}[t]
	\centering
	\resizebox{1.0\textwidth}{!}
	{
		\includegraphics[scale=1.00]{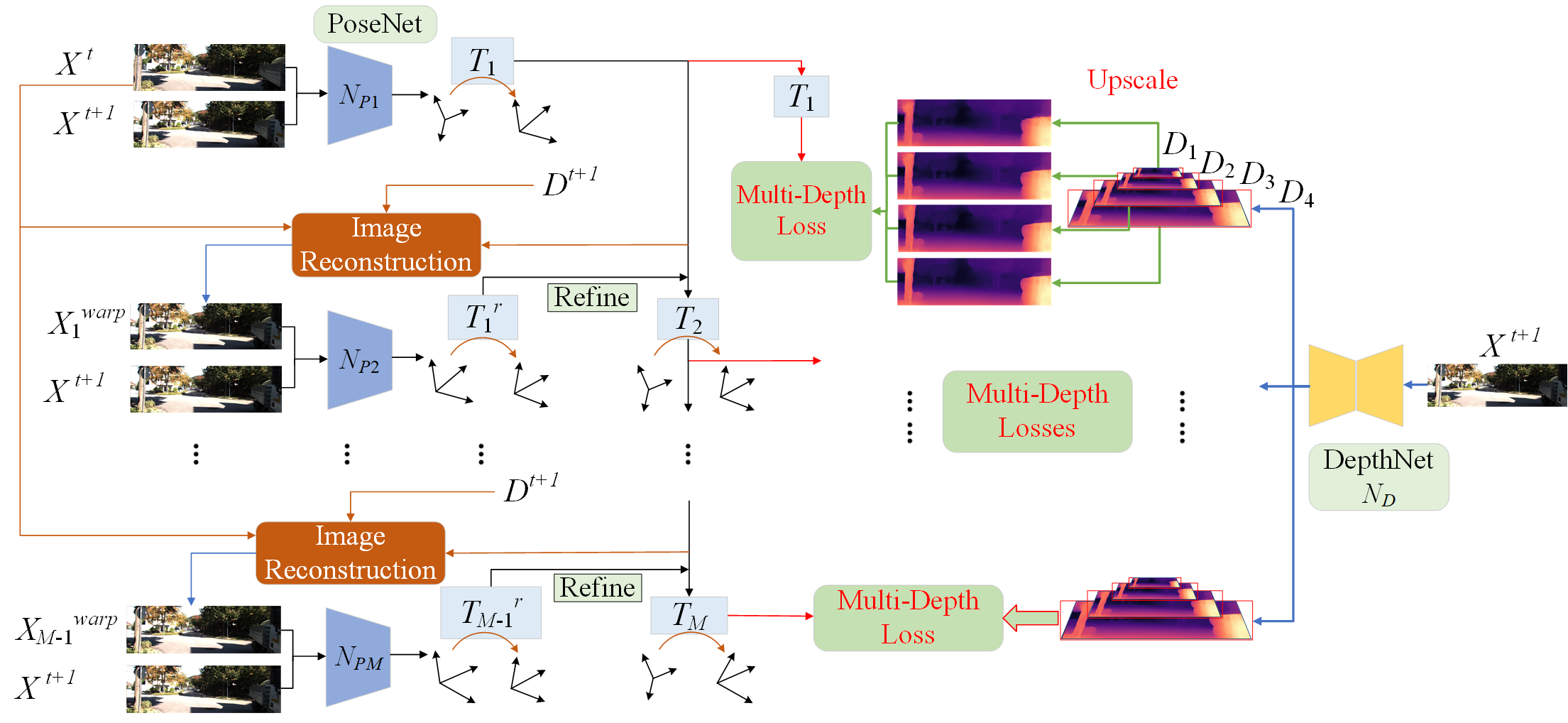}}
	\vspace{-4mm}
	\caption{The details of the pose refinement architecture. The PoseNets are used in series. The first PoseNet $N_{P1}$ estimates a coarse pose from raw input adjacent images. The following PoseNets estimate the residual pose from the raw target images $X^{t+1}$ and the warped images $X^{warp}_{i}$ $(i=1,2,...,M-1)$ reconstructed from the source images $X^{t}$ with the last coarse pose. The coarse pose and the residual poses are combined to obtain the pose in each level.  The DepthNet estimates depth maps at four scales. This figure shows the situation that reconstruction loss is generated by all-scale depth maps and all-level poses. Low-scale depth maps are upsampled to the highest scale as the input to the losses for better training \cite{godard2019digging}.}
	\label{hierarchical_refinement}
\end{figure*}

\subsection{Self-supervised from Monocular Video}

Zhou et al. \cite{zhou2017unsupervised} firstly propose a self-supervised method for depth together with pose estimation by minimizing the pixel-based loss only from monocular video. 
Many studies for self-supervised learning of monocular depth and pose are based on eliminating the influence of the occlusion, dynamics, and illumination variation when reconstructing the image between adjacent frames. To solve these problems, GeoNet \cite{Yin_2018_CVPR} proposes jointly learning for depth, pose and optical flow and use forward optical flow and backward  optical flow to mask occlusion areas. DF-Net \cite{Zou_2018_ECCV} uses depth and pose to get static optical flow, then uses geometry consistency to mask occlusion with estimated static optical flow. They also use the optical flow estimated by the optical flow network and the static flow obtained by the depth-pose networks to train the depth, pose and optical flow tasks. CC \cite{Ranjan_2019_CVPR} and EPC++ \cite{luo2019every} train a motion segmentation net to split static areas for depth and pose learning and dynamic areas for optical flow learning.  Monodepth2~\cite{godard2019digging} uses the minimum photometric error in multiple source frames for each pixel, which most likely does not contain the occlusion and dynamics information. SC-SFM \cite{bian2019depth} masks occluded pixels in images by computing the depth inconsistency probability map and proposes the scale consistency.
Gordon et al.~\cite{Gordon_2019_ICCV} learn the camera's parameters from the video to make the depth prediction more accurate, and use the predicted depth map directly to deal with occlusion.
DOP \cite{wang2020unsupervised} divides an image into three regions, static regions, dynamic regions, and occlusion regions by the information of adjacent frames. They solve the occlusion problem by explicit geometric calculation using the predicted point cloud. D3VO \cite{Yang_2020_CVPR} predicts transform parameters to align the brightness condition to reduce the impact of illumination variation. LEGO \cite{yang2018lego} estimates edge and 3D information together to improve the accuracy of detail estimation. 
Shen et al. \cite{shen2019beyond} introduce geometric constraint matching loss to compensate for the limitation of unsupervised loss of depth and pose estimation.
Mahjourian et al. \cite{mahjourian2018unsupervised} construct the 3D point clouds from estimated depth maps and propose a new 3d Iterative Closest Point (ICP) loss.
Struct2depth \cite{casser2019depth} estimates the motion of each dynamic object in the scene respectively to improve the accuracy of depth and pose estimation in the dynamic scene, and proposes an online refinement method to improve the model's adaptability to new environments.

A better network structure can improve the performance of the depth and pose estimation. 
CM-VO \cite{9345430} calculates the confidence of poses and refine them. GANVO~\cite{GANVO} uses a generator and discriminator structure to optimize the reconstructed image. 
Wang et al. \cite{wang2018learning} propose a differentiable implementation of DVO (DDVO module) that takes full advantage of the relationship between camera pose and depth prediction.
Zhao et al. \cite{zhao2019geometry} use Generative Adversarial Networks (GAN) to generate real images and virtual images, and propose the depth consistency loss between the depth estimated from virtual image and  the depth estimated from real image, which improves the performance of depth estimation network.
Li et al. \cite{li2020unsupervised} propose a new residual translation field regularization method. They decompose the translation field into background translation and object translation with respect to background. 
DPSNet \cite{zhang2021dpsnet} proposes a semantic segmentation method based on geometric reasoning, which performs depth and camera pose estimation and semantic scene segmentation jointly.

Different from above, we focus on a new pose refinement method based on end-to-end fashion and 6-DoF pose augmentation to realize robustness and superiority. To the best of our knowledge, This paper is the first to consider and realize 6-DoF pose augmentation for visual odometry.

\section{Main Approach}
\subsection{Method Overview}
Our method aims to promote the unsupervised learning performance of monocular depth and pose only from monocular video. As the pose estimation is critical to depth training, we propose the 3D hierarchical refinement of pose to improve the depth and pose estimation performance. In the proposed 3D hierarchical refinement architecture, DepthNet $N_D$ is used twice for two monocular input images, $X^t$ and $X^{t+1}$, to estimate the monocular depth maps,  $D^t$ and $D^{t+1}$, respectively, in time sequences. The PoseNets $N_P$ are used in series. The pose $T_{1}$ is estimated by the first PoseNet $N_{P1}$. The following PoseNets $N_{P2}$, $N_{P3}$... estimate the residual pose transformation to realize the hierarchical refinement of the estimated pose. However, the pose transformation is for 3D points, and the input images to the PoseNets are 2D data. Therefore, the transformed 2D images are required as the intermediate state to realize the residual pose learning. The joint unsupervised learning of depth and pose provides this condition. The estimated depth is used based on 2D-3D transformation to obtain the hierarchical intermediate state. 
Through the image reconstruction loss ${L^R}$, geometry consistency loss $L^{GC}$, and depth smooth loss $L^{smooth}$, the multi-scale depths and multi-layer poses are trained together. The three losses will be described detailly in Sec. \ref{multi}. As the KITTI dataset \cite{doi:10.1177/0278364913491297} has an unbalanced and fixed pose range \cite{wang2020tartanair}, in order to expand the data distribution cheaply, we propose the data augmentation loss $L^{aug}$ based on 2D-3D transformation in Sec. \ref{augimg}.

The overall training loss is: 
\begin{equation}
L = \alpha_1  {L^R} + \alpha_2  {L^{GC}} + \alpha_3 L^{smooth}+\alpha_4 L^{aug},\label{eqn:loss}
\end{equation}
where $\alpha_1$, $\alpha_2$, $\alpha_3$, and $\alpha_4$ are hyperparameters.

\begin{figure}[t]
	\centering
	\resizebox{0.95\columnwidth}{!}
	{
		\includegraphics[scale=1.00]{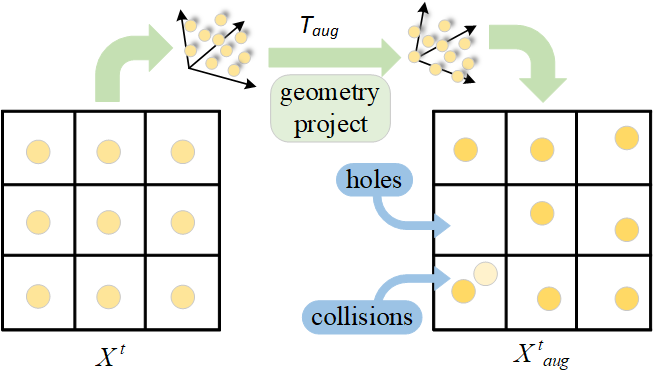}}
	\vspace{-4mm}
	\caption{The details of forward warping. Given a random pose $T_{aug}$, pixels in the original images $X^t$ are projected to new pixel positions on the augmented images $X^t_{aug}$. The figure shows two notable issues in forward warping, i.e. holes and collisions.}
	\label{augwarp}
\end{figure}

\begin{figure*}[ht]
\centering
\subfigure[Raw image]{
\includegraphics[width=5.5cm]{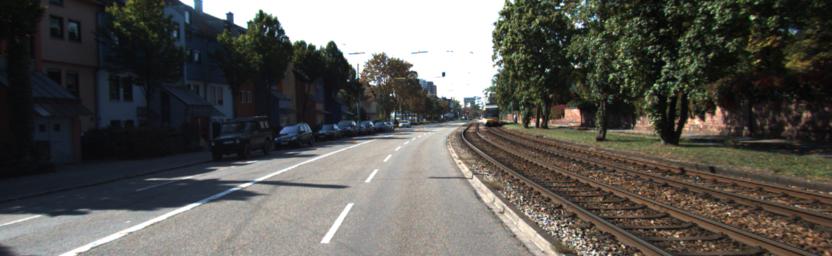}
}
\subfigure[Augmented raw image]{
\includegraphics[width=5.5cm]{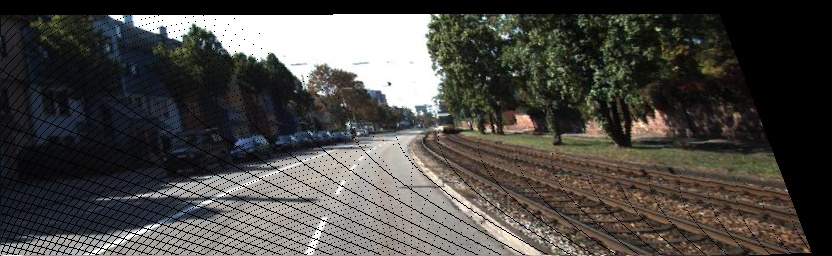}
}
\subfigure[Binary mask $H'$]{
\includegraphics[width=5.5cm]{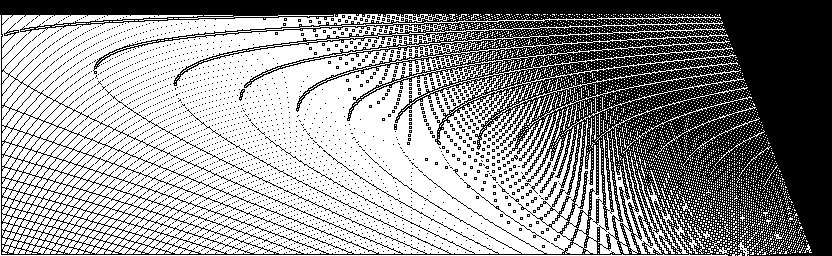} \label{H'}
}
\subfigure[Dilated mask $H_2$]{
\includegraphics[width=5.5cm]{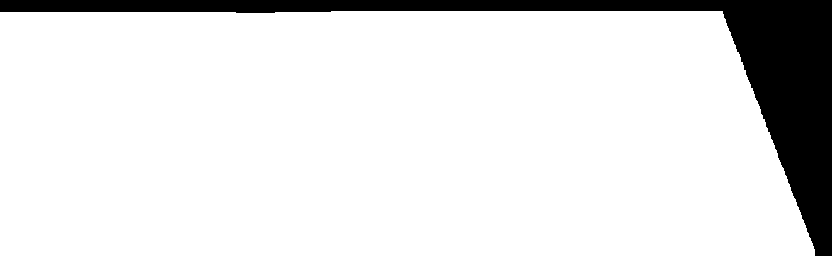}
}
\subfigure[Inpainting mask $H_3$]{
\includegraphics[width=5.5cm]{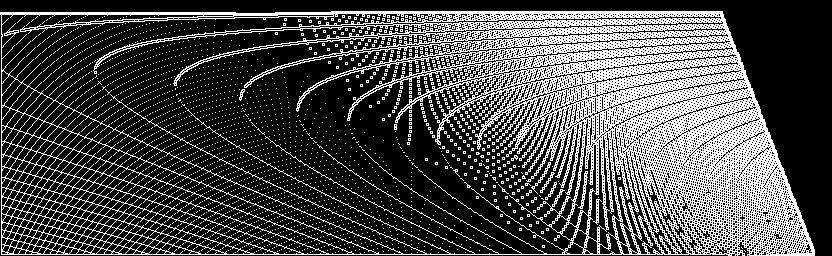}
}
\subfigure[Final augmented image]{
\includegraphics[width=5.5cm]{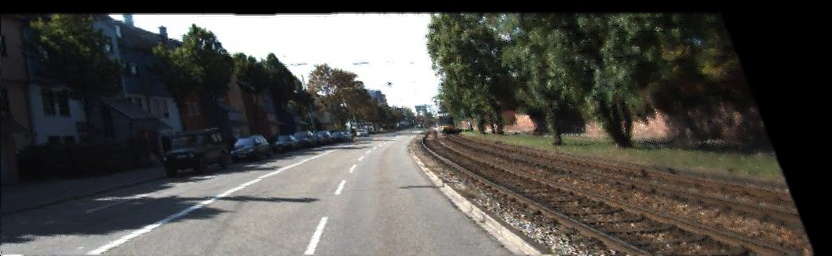}
}
\caption{An example of our hole filling method: (a) is the raw image. (b) is the raw augmented image generated by (a) using forward warp. (c) is the binary mask $H'$. (d) is the dilated mask $H_2$. (e) is the inpainting mask $H_3$. (f) is our final augmented image.} \label{hole_filling}
\end{figure*}

\begin{figure*}[t]
	\centering
	\resizebox{0.85\textwidth}{!}
	{
		\includegraphics[scale=1.00]{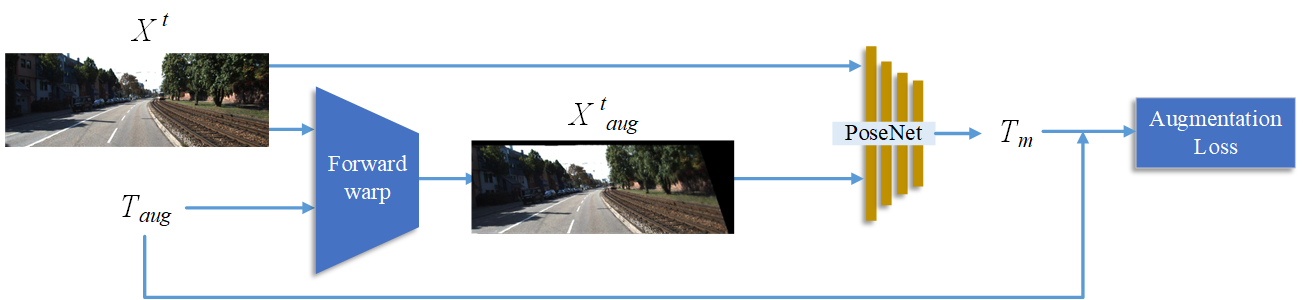}}
	\caption{The details of pose augmentation learning in Sec. \ref{augimg}. The original image $X^t$ and the random pose $T_{aug}$ are input to the process of forward warp to generate augmented image $X^t_{aug}$. Then, $T_m$ is obtained by inputting $X^t$ and $X^t_{aug}$ to the PoseNet. Finally, we get the augmentation loss by formula (\ref{equ:AugLoss}).}
	\label{augmentation}
\end{figure*}

\subsection{3D Hierarchical Refinement with Multi-scale Depths and Multiple Poses} \label{multi}
Our DepthNet $N_D$ has multi-scale depth estimation results, $D_1, D_2, ..., D_{N}$, where $D_1$ means the minimum scale of depth map.
Our System has multiple PoseNets, $N_{P1}, N_{P2},...,N_{PM}$, with the same network structure to refine the pose estimation. The idea is introduced in Fig. \ref{origin}. $X^t$ is obtained when the robot is at $P^t$, while $X^{t+1}$ is obtained at $P^{t+1}$. The first PoseNet $N_{P1}$ estimate the  coarsest pose $T_1$. Then, we use the image warping based on 2D-3D transformation to construct the intermediate image between $X^t$ and $X^{t+1}$ to make next residual pose estimation easier. 
We construct an image $X_{1}^{warp}$ of a virtual intermediate view based on the estimated coarsest pose $T_1$ as following 2D-3D transformation:
\begin{equation}
\begin{aligned}
	& D_{1}^{pro}(i_{pro},j_{pro})\cdot[{i_{pro}},{j_{pro}},1] = \\
	& K\cdot{T_1}\cdot(D^{t + 1}(i^{t+1},j^{t+1}) \cdot {K^{ - 1}} \cdot {[{i^{t + 1}},{j^{t+ 1}},1]^T}),\label{eqn:pro}
\end{aligned}
\end{equation}
\begin{equation}
	X_{1}^{warp}({i^{t + 1}},{j^{t + 1}}) = \sum\limits_{i \in \{ t,b\} ,j \in \{ l.r\} } {{w^{ij}}{X^t}(i,j)},\label{eqn:warp}
\end{equation}
where $K$ is the intrinsic matrix. $({i_{pro}},{j_{pro}})$ is the projected coordinates in $X_t$ image plane from $({i^{t+1}},{j^{t+1}})$. $(t,l),(t,r),(b,l),(b,r)$ are four adjacent pixel coordinates around $({i_{pro}},{j_{pro}})$, and the sum of $w^{ij}$ for the four adjacent pixel equals $1$.
That is, $X_{1}^{warp}$ is a virtual image perceived at the view of $P_1^{warp} = T_1 \times P^t$, which is a closer view to $P^{t+1}$ than camera view  $P^{t}$as shown in Fig. \ref{origin}. 
As shown in Fig. \ref{hierarchical_refinement}, new $X_{1}^{warp}$ and $X^{t+1}$ are input to the second PoseNet $N_{P2}$ and obtain the residual pose estimation $T_1^r$. The refined pose estimation between raw two adjacent images can be calculated by:
\begin{equation}
	T_2= T_1^r \cdot T_1 = N_{P2}(X_{1}^{warp}, X^{t+1}) \cdot T_1. \label{eqn:t}
\end{equation}
Through the refined pose, the new intermediate view picture can be synthesized and input to the next PoseNet combined with $X^{t+1}$ for the next pose refining. All these calculations are differentiable so that we can optimize the whole system end-to-end.

Finally, our system has multi-layer pose estimation results, $T_1, T_2, ..., T_m, ..., T_{M}$, where $T_1$ means the coarsest pose estimation result. The depth estimation results $D_1, D_2, ..., D_n, ..., D_{N}$ are multi-scale, where $D_1$ is the lowest scale. So that, there are a number of combinations of the depths and poses to calculate $X_{1}^{warp}$ in the formula (\ref{eqn:pro}) and (\ref{eqn:warp}). $X_{m,n}^{warp}$ denotes the warped image obtained by the formula (\ref{eqn:pro}) and (\ref{eqn:warp}) with $T_{m}$ and $D_{n}$. The warped image $X_{m,n}^{warp}$ and raw $X^{t+1}$ are used to calculated image reconstruction loss $\sigma (X_{m,n}^{warp},{X^{t + 1}})$ to constrain the training of depth and pose networks to realize the unsupervised learning of depth and pose:
\begin{equation}
	\begin{gathered}
		\begin{split}
			\sigma (X_{m,n}^{warp},{X^{t + 1}}) =& {\lambda _\rho }{\left\| {X_{m,n}^{warp} - {X^{t + 1}}} \right\|_1} \hfill \\
			+& (1 - {\lambda _\rho })\frac{{1 - SSIM(X_{m,n}^{warp},{X^{t + 1}})}}{2}, \hfill \\ 
		\end{split} 
	\end{gathered} \label{lambda}
\end{equation}
where $\lambda _\rho$ is a hyperparameter. $SSIM(X_{m,n}^{warp},{X^{t + 1}})$ is the the structural similarity (SSIM) \cite{wang2004image} widely used in \cite{GANVO, 8793622, bian2019depth, godard2019digging}. However, due to occlusion, there are some pixels can not be constructed from adjacent frame, which produce some vacancy in $X_{m,n}^{warp}$. 
The depth scale consistence \cite{bian2019depth} is used to obtain the occlusion mask. Specifically, the depth inconsistency map is calculated by:
\begin{equation}
D_{m,n}^{diff}({i^{t + 1}},{j^{t + 1}}) = \frac{{\left| {D_{m,n}^{pro}({i^{t + 1}},{j^{t + 1}}) - D_{m,n}^{warp}({i^{t + 1}},{j^{t + 1}})} \right|}}{{D_{m,n}^{pro}({i^{t + 1}},{j^{t + 1}}) + D_{m,n}^{warp}({i^{t + 1}},{j^{t + 1}})}}, \label{eqn:diff}
\end{equation}
Then, the weight mask can be obtained by:
\begin{equation}M_{m,n}^{weight}({i^{t + 1}},{j^{t + 1}}) = 1- D_{m,n}^{diff}({i^{t + 1}},{j^{t + 1}}), \label{eqn:weight}\end{equation}
where ${D_{m,n}^{pro}}$ is obtained from formula (\ref{eqn:pro}). $D_{m,n}^{warp}$ is obtained by bilinear interpolation like $X_{m,n}^{warp}$ in formula (\ref{eqn:warp}).
The auto-mask \cite{godard2019digging} is also used to filter the objects which are static relative to the camera or textureless regions, because these relatively static objects have different motion patterns with the overall pose estimation, and textureless regions make the networks confused about the pixel match for these regions. The auto-mask is calculated as follows:
\begin{equation}{M_{m,n}^{auto}} = \sigma (X_{m,n}^{warp},{X^{t + 1}}) < \sigma ({X^{t}},{X^{t + 1}}).\end{equation}
The final image reconstruction loss is:
\begin{equation}
	{L_{m,n}^R} = \frac{1}{{\left| \Omega  \right|} }\sum\limits_\Omega M_{m,n}^{weight}M_{m,n}^{auto} \sigma (X_{m,n}^{t + 1},{X^{t + 1}}),\label{eqn:l1}\end{equation}
where $\Omega$ means all pixels for an image and ${\left| \Omega  \right|}$ means the amount of pixels in an image.  Scale consistency loss \cite{bian2019depth} is used to learn the consistent scale for the depth and pose in consecutive frames:
\begin{equation}
L_{m,n}^{GC} = \frac{1}{{\left| \Omega  \right|} }\sum\limits_\Omega M_{m,n}^{auto} {{D_{m,n}^{diff}}}. 
\label{eqn:lmngc}
\end{equation}
$D_{m,n}^{diff}$ is obtained from formula (\ref{eqn:diff}), which measures the difference of depth estimation between two adjacent frames. Therefore, formula (\ref{eqn:lmngc})  encourage the scale consistency in two adjacent frames. As the last refined pose $P_{M}$ is most accurate in the estimated multi-layer poses, $P_{M}$ is used to optimize all depths. However, coarser poses can have a negative effect on depth estimation in joint training. Therefore, stopping gradient is made for all scale depths, $D_1,D_2,...,D_{N}$, when they are optimized with the coarser pose, $T_1,T_2,...,T_{M-1}$. 
The overall image reconstruction loss and geometry consistency loss are:
\begin{equation}
		{L^R} = stop\_D(\sum\limits_{\begin{subarray}{l} 
			u \in \{ 1,2,...,M - 1\} , \\ 
			v \in \{ 1,2,...,N\} , 
		\end{subarray}}  {L_{u,v}^R} ) + \sum\limits_{v \in \{ 1,2,...,N\} } {L_{M,v}^R},
\end{equation}
\begin{equation}
	\begin{gathered}
		{L^{GC}} = stop\_D(\sum\limits_{\begin{subarray}{l} 
			u \in \{ 1,2,...,M - 1\} , \\ 
			v \in \{ 1,2,...,N\} , 
		\end{subarray}}  {L_{u,v}^{GC}} ) + \sum\limits_{v \in \{ 1,2,...,N\} } {L_{M,v}^{GC}},
	\end{gathered} 
\end{equation}
where $stop\_D()$ means stopping gradient for the input depth. The ablation study in Sec. \ref{ablation} will demonstrate the effectiveness of this design.

The edge-aware depth smoothness loss is used to make the estimated depth smoother inside an object but remains unsmooth at the edge of the object. The revised depth smooth loss \cite{wang2020unsupervised} is used to mitigate depth degradation in the training process:
\begin{equation}{L^{smooth}} = \frac{1}{\Omega }\sum\limits_\Omega  {\left\| {\nabla (\frac{{{D^{t + 1}}}}{{\min ({D^{t + 1}})}}) \cdot {e^{ - \nabla {X^{t + 1}}}}} \right\|}.  \end{equation}
For the convenience of explanation, only losses for $t+1$ are presented above. By projection in the opposite direction, losses for $t$ are also used.

\subsection{6-DoF Pose Augmentation by Means of Forward Warping}  \label{augimg}

\setlength{\tabcolsep}{1.5mm}
\begin{table*}[t]
	\caption{Depth evaluation results using Eigen et al. test split \cite{eigen2014depth} on KITTI dataset. Previous supervised and unsupervised monocular depth learning methods are compared with ours. \textbf{D}: Depth supervision. \textbf{S}: self-supervised from stereo video. \textbf{M}: self-supervised from monocular video. \textbf{AT}: auxiliary task.}
	\footnotesize
	\begin{center}
		\resizebox{0.9\textwidth}{!}
		{
			\begin{tabular}{l|c|c|cccc|ccc}
				\hline
				\multirow{2}{*}{Methods}&  \multirow{2}{*}{with AT}&\multirow{2}{*}{Train}& AbsRel & SqRel & RMSE & RMSE log & $\delta$ < 1.25 & $\delta < 1.25^2$ &  $\delta < 1.25^2$\\
				\cline{4-10}
				&  & & \multicolumn{4}{c|}{Lower is better} & \multicolumn{3}{c}{Higher is better}\\
				\hline
				Eigen et al.~\cite{eigen2014depth} && D 
				& 0.203 & 1.548 & 6.307 & 0.282 & 0.702 & 0.890 &  0.958\\
				Liu at al. ~\cite{liu2015learning} && D 
				& 0.202 & 1.614 & 6.523 & 0.275 & 0.678 & 0.895 &  0.965\\
				Garg et al. ~\cite{garg2016unsupervised} &&S& 0.152 &1.226 &5.849 &0.246 &0.784 &0.921 &0.967\\
				
				Godard et al. \cite{godard2017unsupervised} & &S& 0.148 &1.344& 5.927 &0.247 &0.803& 0.922 &0.964 \\
				Zhan et al. \cite{zhan2018unsupervised} &&S& 0.144 &1.391& 5.869& 0.241 &0.803 &0.928& 0.969\\ 
				Kuznietsov et al. ~\cite{Kuznietsov_2017_CVPR} && DS 
				& 0.113 & 0.741 & 4.621 & 0.189 & 0.862&  0.960 & 0.986\\
				Guo et al.~\cite{Guo_2018_ECCV} & &DS 
				& 0.096 & 0.641 & \bf4.095 & \bf0.168 & 0.892 & 0.967 & 0.986\\
				Tian et al.~\cite{tian2021depth} & &S 
				& \bf0.095 & \bf0.601 & 4.128 & 0.176 & \bf0.908 & \bf0.976 & \bf0.991\\
				\hline
				Zhou et al.~\cite{zhou2017unsupervised} && M 
				& 0.208 & 1.768 & 6.856 & 0.283 & 0.678 & 0.885 & 0.957\\
				Yang et al.~\cite{yang2018unsupervised} &$\checkmark$ & M 
				& 0.182 & 1.481 & 6.501 & 0.267 & 0.725 & 0.906 & 0.963\\
				Mahjourian et al. \cite{mahjourian2018unsupervised} && M& 0.163 &1.240 &6.220 &0.250 &0.762 &0.916 &0.968\\
				 LEGO \cite{yang2018lego} & & M & 0.162 & 1.352 & 6.276 & 0.252 & - & - &-\\
				Geonet-VGG~\cite{Yin_2018_CVPR} &$\checkmark$ &M 
				& 0.164 & 1.303 & 6.090 & 0.247 & 0.765 & 0.919 & 0.968\\
				Geonet-Resnet~\cite{Yin_2018_CVPR} &$\checkmark$ & M 
				& 0.155 & 1.296 & 5.857 & 0.233 & 0.793 & 0.931 & 0.973\\
				Shen et al. \cite{shen2019beyond} & & M & 0.156 & 1.309 & 5.37 &0.236 & 0.797 & 0.929 & 0.969 \\
				DPSNet \cite{zhang2021dpsnet} & $\checkmark$ & M &0.159 & 1.355& 5.679 & 0.232 & 0.785 & 0.935 &0.973\\
				Wang et al. \cite{wang2018learning} &&M& 0.151 &1.257& 5.583 &0.228& 0.810 &0.936 &0.974\\
				DF-Net~\cite{Zou_2018_ECCV} &$\checkmark$ & M 
				& 0.150 & 1.124 & 5.507 & 0.223 & 0.806 & 0.933 & 0.973\\
				GANVO~\cite{GANVO} && M 
				& 0.150 & 1.141 & 5.448 & 0.216 & 0.808 & 0.939 & 0.975\\
				CC~\cite{Ranjan_2019_CVPR} &$\checkmark$ &M 
				& 0.140  & 1.070  & 5.326  & 0.217  & 0.826  & 0.941  & 0.975\\
				EPC++ \cite{luo2019every} &$\checkmark$ &M& 0.141& 1.029 &5.350 &0.216 &0.816 &0.941 &0.976\\
				Struct2depth \cite{casser2019depth} & & M & 0.141 & 1.026 & 5.291 & 0.215 & 0.816 & 0.945 & 0.979 \\
				DOP~\cite{wang2020unsupervised} &$\checkmark$ & M 
				& 0.140 & 1.068 & 5.255 & 0.217 & 0.827 & 0.943 & 0.977\\
				SC-SFM~\cite{bian2019depth}&& M 
				& 0.137 & 1.089 & 5.439 & 0.217 & 0.830 & 0.942 & 0.975\\
				Gordon et al.~\cite{Gordon_2019_ICCV} && M 
				& 0.128 & 0.959 & 5.230 & -  & - & - & -\\
				Monodepth2~\cite{godard2019digging} & &M 
				& 0.115 &0.882 &4.701 &0.190 &0.879 &0.961 &0.982\\
				Li et al. \cite{li2020unsupervised} &&M& 0.130& 0.950& 5.138& 0.209 &0.843 &0.948 &0.978\\
				$\Omega$Net \cite{tosi2020distilled} &$\checkmark$ &M
				& 0.125 &0.805 &4.795 &0.195 &0.849 &0.955 &\bf0.983\\
				Chen et al. \cite{chen2021fixing}  & & M&0.129 &0.976 & 4.958 & 0.203 & 8.848 & 0.951 & 0.979\\
				Jia et al. \cite{jia2021self}& &M &0.130 & 0.957 &4.907 & 0.203 & 0.851 & 0.954 & 0.980 \\ 
				
				Ours && M 
				& \bf0.109 &\bf0.790 &\bf4.656 &\bf0.185 &\bf0.882 &\bf0.962 &\bf0.983\\
				\hline
		\end{tabular}}
	\end{center}
	\label{depth}
\end{table*}

As the pose distribution of the KITTI dataset \cite{doi:10.1177/0278364913491297} is not uniform,  a virtual dataset that includes an extensive distribution of pose data is proposed by Wang et al. \cite{wang2020tartanair}. However, the data gap is inevitable from the virtual data to the real autonomous driving scenario. Therefore, we propose a 6-DoF pose augmentation by generating a random pose $T_{aug}$ and then forward warping \cite{10.1007/978-3-030-58452-8_42} the origin image $X^t$ to synthesizing a new arbitrary view $X_{aug}^t$ :
\begin{equation}
\begin{aligned}
	& D_{pro}^{aug}(i_{pro},j_{pro})\cdot[{i_{pro}},{j_{pro}},1] =\\
	& K\cdot{T_{aug}}\cdot(D^{t}(i^t,j^t)\cdot {K^{ - 1}} \cdot {[{i^{t}},{j^{t}},1]^T}),\label{eqn:fpro}
\end{aligned}
\end{equation}
where $K$ is the intrinsic matrix. $({i^t},{j^t})$, $({i_{pro}},{j_{pro}})$ are respectively the pixel coordinates in the image plane of $X^t$ and  $X^{t}_{aug}$. $D^t(i^t,j^t)$ and $D^{aug}_{pro}(i_{pro},j_{pro})$ are respectively the depth of $({i^t},{j^t})$ and $({i_{pro}},{j_{pro}})$. The details of forward warping is shown in Fig. \ref{augwarp}.

Due to occlusion in the transformation, collisions may occur, that is, more than one pixel in the original images may be projected to the same grid in the augmented images. To solve the collision problem,  pixels with minimum depth in original images $X^t$ are selected to be displayed on the augmented image $X^{aug}$. Apart from collisions, the occurrence of holes is another notable issue in forward warping, which means that pixels of the augmented image $X^{aug}$ may have no projection points from the original image $X^t$. To handle holes, a novel hole filling method is established.

\begin{figure*}[h]
	\centering
	\begin{tabular}{p{4.3cm}<{\centering}p{4.1cm}<{\centering}p{4.1cm}<{\centering}p{4.3cm}<{\centering}}
		Input &  SC-SFM~\cite{bian2019depth} & Monodepth2~\cite{godard2019digging}  & Ours
		\vspace{-0.5cm}
	\end{tabular}
	\includegraphics[width=1.0\textwidth]{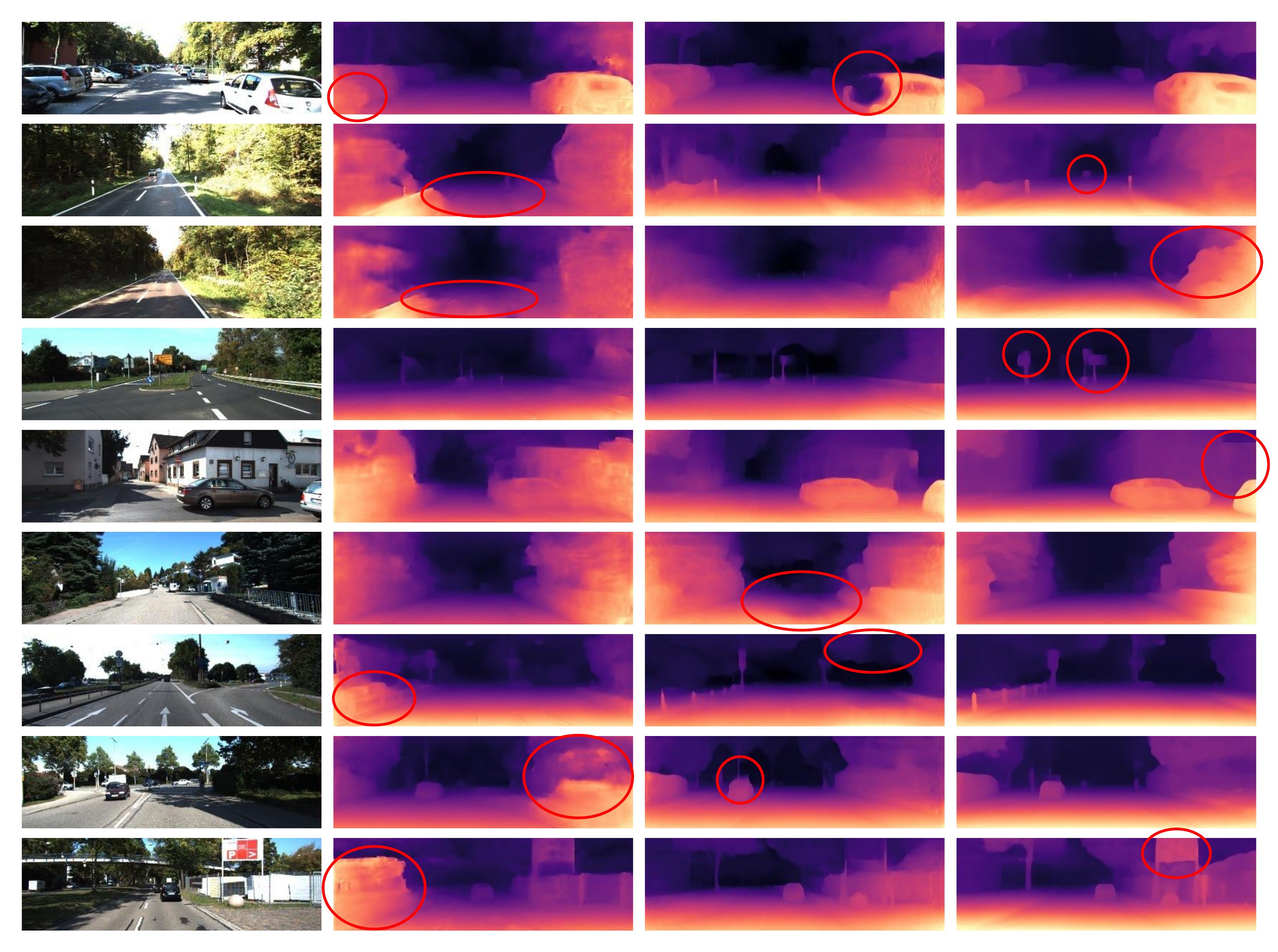}
	\vspace{-0.9cm}
	\caption{Qualitative results for depth estimation on the KITTI Eigen test split \cite{eigen2014depth}. Our depth estimation model provides depth maps with higher quality, as shown in the last column.}
	\label{depth_visual}
\end{figure*}

6-DoF pose augmentation vastly increases the chances of holes. To remove these holes, we first get a binary mask $H'$ referred to \cite{Aleotti_2021_CVPR}, where pixels with holes in augmented images are labeled as 0 and others as 1. Then, an inpainting strategy \cite{telea2004image} is utilized to fill the holes by applying the binary mask $H'$. However, we find that there is a large area of continuous holes in the augmented image as shown in Fig. \ref{H'}, leading the inpainting strategy not accurate enough. Thus, we propose a novel hole filling method to handle this problem, the main idea of which is to fill the small area of holes by employing inpainting strategy and fill continuous holes with zero. Specifically, we first dilate mask $H'$ to $H_2$, and then get the binary inpainting mask $H_3$ by subtracting $H_2$ and $H'$:
\begin{equation}
    H_3= H_2- H'.
\end{equation}
Finally, we can apply the inpainting strategy only to pixels labeled with 1 in $H_3$. An example of our hole filling method is shown in Fig. \ref{hole_filling}.

The augmentation pose is random but known between the original image and the augmented image. This property is used to train the PoseNets by supervised training. The original image and the augmented image are input to the PoseNets as shown in Fig. \ref{augmentation}. The output of the PoseNets is $T_m$. The pose-supervised loss is used for this training:
\begin{equation}
\begin{split}
    L^{pose}_{aug} = \| t_{m}-t_{aug}\|_{2} \exp(-w_t) \\
	+w_t+\| q_{m}-q_{aug}\|_{2} \exp(-w_q)+w_q,
\end{split}
\label{equ:AugLoss}
\end{equation}
where $\|\cdot \|$ denotes the $l_2$-norm. $t_{aug}$ and $q_{aug}$ are respectively translation vector and quaternion generated by known random pose $T_{aug}$. $t_{m}$ and $q_{m}$ are respectively translation vector and quaternion generated by the pose $T_{m}$. $w_t$ and $w_q$ are learnable weights. 

\begin{table}[h]
	\caption{Visual odometry results on sequences 09 and 10 of KITTI odometry dataset~\cite{doi:10.1177/0278364913491297}. 
		\textbf{AT}: auxiliary task. \textbf{LC}: loop closure.}
	\footnotesize
	\begin{center}
		\resizebox{0.9\columnwidth}{!}
		{
			\begin{tabular}{l|cc|cc}
				\hline
				\multirow{2}{*}{Methods} & \multicolumn{2}{c|}{seq.09} & \multicolumn{2}{c}{seq.10} \\
				\cline{2-5}
				& $t_{rel}$ & $r_{rel}$ & $t_{rel}$ & $r_{rel}$\\
				\hline
				ORB-SLAM2 ~\cite{mur2017orb} &9.31 &0.26 &2.66 &0.39\\
				ORB-SLAM2 (LC)~\cite{mur2017orb} & \bf2.84 &\bf0.25 &\bf2.67 &\bf0.38\\
				\hline
				Zhou et al.~\cite{zhou2017unsupervised} & 17.84 & 6.78 & 37.91 & 17.78\\
				Zhan et al. ~\cite{zhan2018unsupervised} & 11.93 & 3.91 & 12.45 & 3.46\\
				SC-SFM ~\cite{bian2019depth} & 11.2 & 3.35 & 10.1 & 4.96\\
				CM-VO ~\cite{9345430} & 9.69 & 3.37 & 10.01 & 4.87 \\
				Monodepth2~\cite{godard2019digging} & 8.20 & 1.76 & 8.39 & 3.21\\
				CC~\cite{Ranjan_2019_CVPR} (with AT) & 7.71 & 2.32 & 9.87 & 4.47 \\
				DOP~\cite{wang2020unsupervised} (with AT) & 5.65 & 1.81 & 8.31 & 2.65\\
				Ours & \bf3.37 & \bf0.82 & \bf2.76 & \bf0.90\\
				\hline
		\end{tabular}}
	\end{center}
	\label{pose}
\end{table}

\begin{figure}
	\centering
	\setlength{\tabcolsep}{1.0mm}
	\begin{tabular}{cc}
		\includegraphics[width=0.58\linewidth]{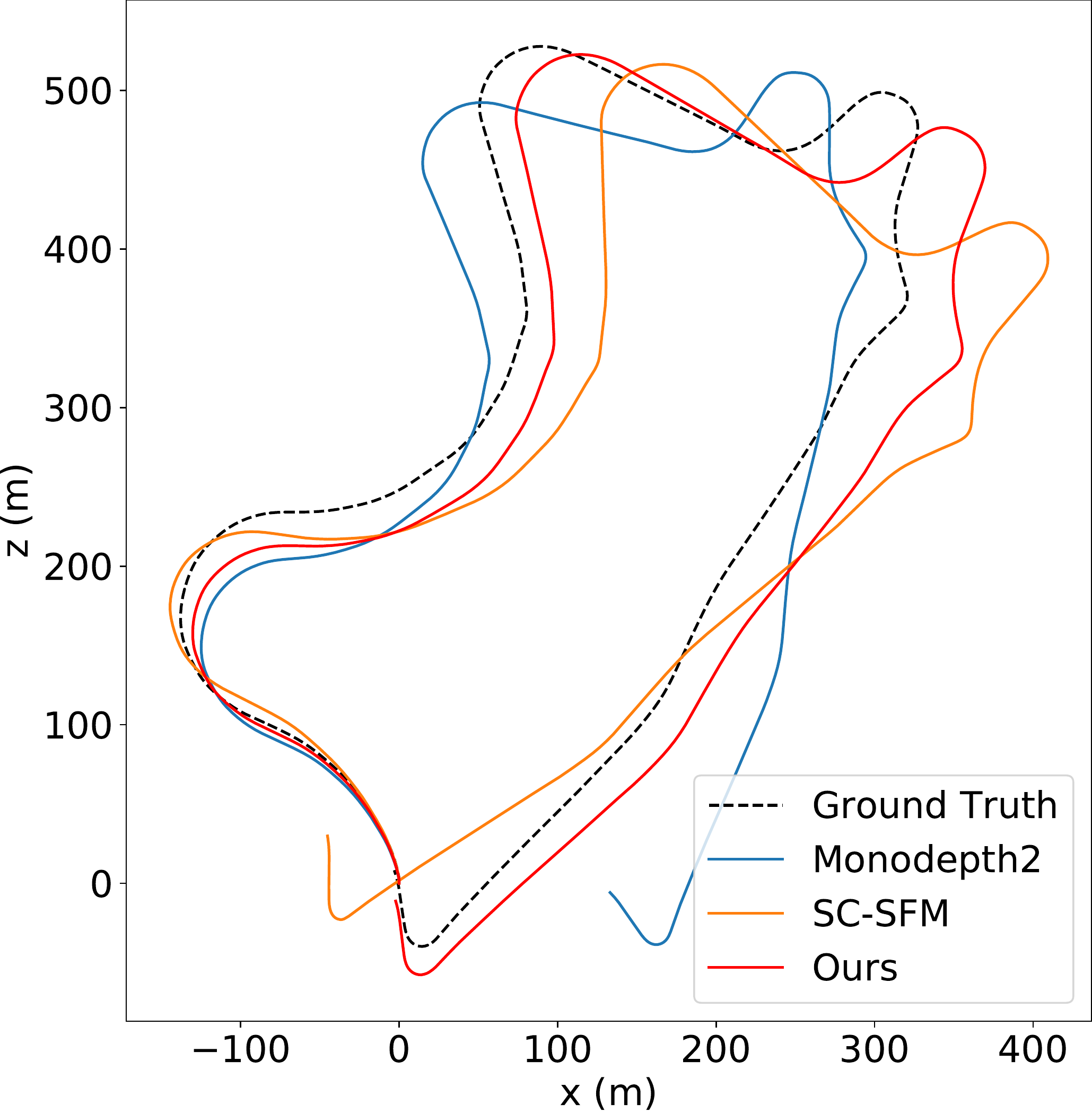} & 
		\includegraphics[width=0.312\linewidth]{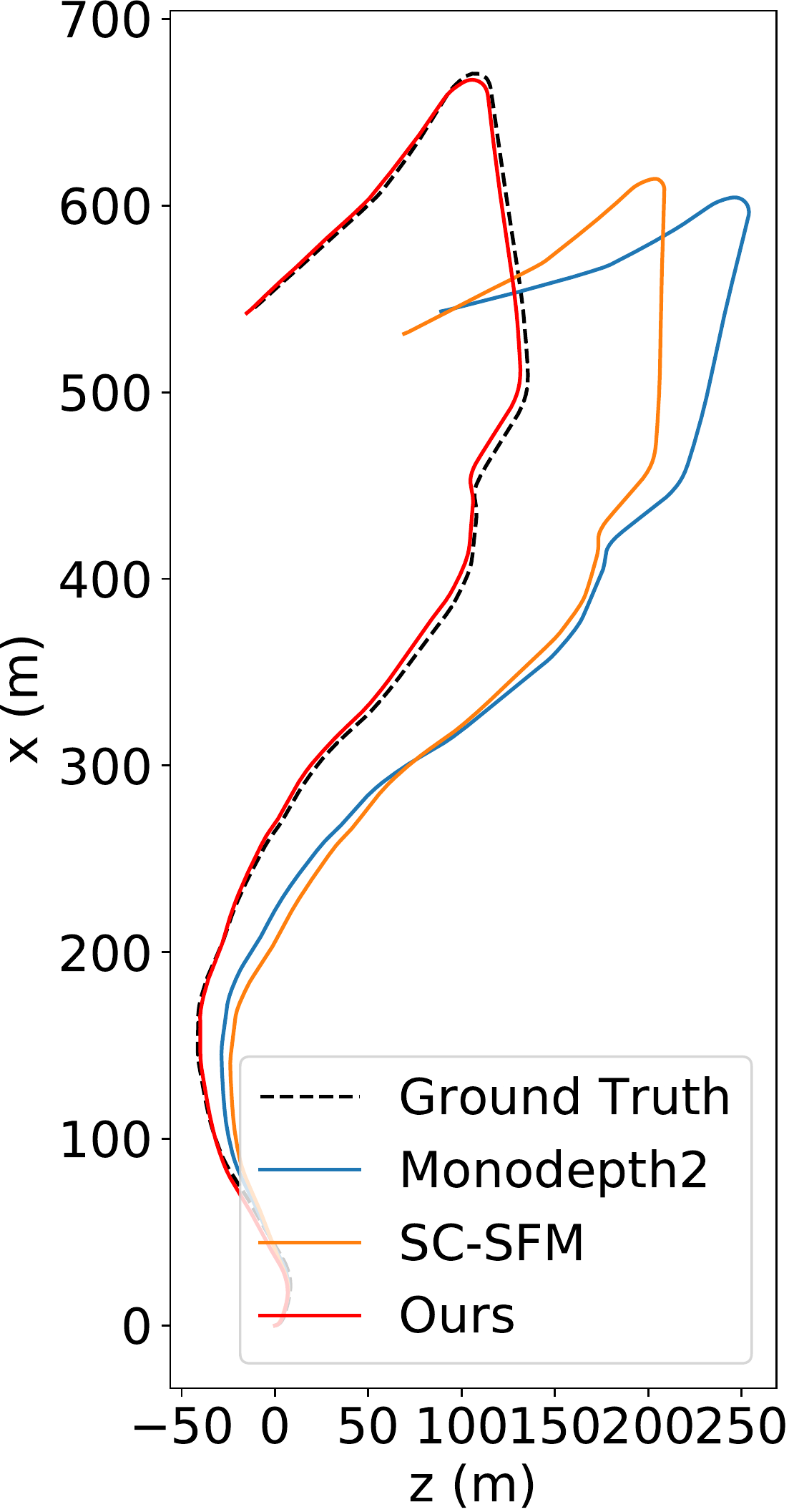} \\
	\end{tabular}
	\caption{Visual odometry results on sequences 09 (left) and 10 (right) of KITTI odometry dataset.}
	\label{odometry}
\end{figure}


\setlength{\tabcolsep}{0.9mm}
\begin{table*}
	\caption{Ablation Results for depth estimation using Eigen et al. test split \cite{eigen2014depth} on KITTI dataset.  
	\textbf{A-depth}: all-scale depth. 
	\textbf{A-pose}: all poses. \textbf{H-pose}: highest-refined pose. \textbf{C-pose}: Coarse poses, which are the poses except for the highest-refined pose. $\pmb{\leftrightarrows}$: Losses and gradient are calculated with two sides. $\pmb{\rightarrow}$: Losses are calculated with two sides, but gradients are only calculated for the right side and stopped for the left side.}
	\footnotesize
	\begin{center}
		\resizebox{0.95\textwidth}{!}
		{
			\begin{tabular}{c|l|cccc|ccc}
				\hline
				& \multirow{2}{*}{Methods} & AbsRel & SqRel & RMSE & RMSE log & $\delta$ < 1.25 & $\delta < 1.25^2$ &  $\delta < 1.25^2$\\
				\cline{3-9}
				& & \multicolumn{4}{c|}{Lower is better} & \multicolumn{3}{c}{Higher is better}\\
				\hline
				\multirow{3}{*}{(a)} 
				& Baseline (Single PoseNet)
				&   0.116  &   0.848  &   4.919  &   0.195  &   0.868  &   0.957  &   0.981  \\
				& A-depth$\leftrightarrows$A-pose
				&    0.113   &	0.815   &	4.786   &	0.190   &	0.873   &	0.959   &	\bf0.982\\
				& A-depth$\leftrightarrows$H-pose, A-depth$\rightarrow$C-pose
				&    \bf0.111   &	\bf0.813   &	\bf4.656   &	\bf0.189   &	\bf0.879   &	\bf0.960   &	\bf0.982\\
				\hline
				\multirow{2}{*}{(b)} 
				& Baseline(A-depth$\leftrightarrows$H-pose, A-depth$\rightarrow$C-pose)
				&    0.111   &	0.813   &	\bf4.656   &	0.189   &	0.879   &	0.960   &	\bf0.982\\
				& Pose augmentation&    \bf0.110   &	\bf0.810   &	4.709   &	\bf0.187   &	\bf0.881   &	\bf0.962   &	\bf0.982\\
				\hline
				\multirow{3}{*}{(c)} 
				& Baseline (2-pose, with pose augmentation) &    0.110   &	0.810   &	4.709   &	0.187   &	0.881   &	\bf0.962   &	0.982\\
				
				& 3-pose &    \bf0.109 &0.834 &4.692 &0.186 &\bf0.883 &\bf0.962 &0.982\\
				
				& 4-pose & \bf0.109 &\bf0.790 &\bf4.656 &\bf0.185 &0.882 &\bf0.962 &\bf0.983\\
				\hline
		\end{tabular}}
	\label{depth_ablation}
	\end{center}
\end{table*}

\section{Experiments}\label{exp}
\subsection{Datasets and Implementation Details}\label{detail}
\subsubsection{KITTI Raw Dataset}
The KITTI raw dataset \cite{doi:10.1177/0278364913491297} contains raw color images and depth sensor information collected and divided into the categories ‘Road’, ‘City’, ‘Residential’, ‘Campus’, and ‘Person’. For the evaluation of depth estimation, we follow previous works~\cite{bian2019depth, 9156629} to train and test models on the data split of Eigen et al.~\cite{eigen2014depth} only on KITTI raw dataset~\cite{doi:10.1177/0278364913491297}. 

\subsubsection{KITTI Odometry Dataset}
The KITTI odometry spilits~\cite{doi:10.1177/0278364913491297} contains 11 long driving stereo sequences with available ground truth trajectories. For pose estimation, we evaluate visual odometry results on KITTI odometry dataset~\cite{doi:10.1177/0278364913491297} as previous works ~\cite{bian2019depth,9156629}, using sequences 00-08 for training and sequences 09-10 for testing.

\subsubsection{Implementation Details}
All the experiments except for 3 PoseNets and 4 PoseNets are implemented on a NVIDIA RTX 2080Ti GPU. The experiments about 3 PoseNets and 4 PoseNets are on a NVIDIA Titan RTX GPU. For pose estimation, separate convolution neural networks in ~\cite{godard2019digging} are used to estimate coarse pose and residual poses between two adjacent images. For depth estimation, DispResNet18 architecture~\cite{godard2019digging} is used to generate multi-scale disparity maps from a single image. The image size for all experiments is $832 \times 256$. A snippet of two sequential images is used to estimate both forward pose and backward pose between two frames. The losses are used in two-directional transformation and image reconstruction like SC-SFM ~\cite{bian2019depth}. The hyperparameters are set as  $\alpha_1 = 1.0$, $ \alpha_2 = 0.1$, $ \alpha_3 = 0.5$ and $\alpha_4=2.0$ in formula (\ref{eqn:loss}), $\lambda_{\rho} = 0.15$ in formula (\ref{lambda}). Adam optimizer \cite{kingma2014adam} is used for training, and the learning rate is set as $10^{-4}$. The batch size is 4 for all experiments.

\subsection{Evaluation Results}\label{comparison}

Through the training and test as in Sec. \ref{detail}, our depth estimation results and visual odometry results on the KITTI dataset are shown in Tables \ref{depth} and \ref{pose}, respectively.
On both monocular depth estimation and monocular visual odometry by the unsupervised learning of depth and pose, our method outperforms recent state-of-the-art methods. Besides, our method does not need auxiliary tasks, such as optical flow estimation \cite{Yin_2018_CVPR, Zou_2018_ECCV, Ranjan_2019_CVPR, luo2019every,wang2020unsupervised,tosi2020distilled}, semantic segmentation \cite{tosi2020distilled} or dynamic mask estimation \cite{Ranjan_2019_CVPR, luo2019every}, normal estimation \cite{yang2018unsupervised}. 
This paper realizes end-to-end iterative view synthesis and pose refinement to jointly optimize the pose and depth estimation networks, allowing the overall parameters to self-learn according to the optimization objective. 
We give the qualitative results of our depth estimation compared with other methods, as shown in Fig. \ref{depth_visual}. Our methods perform better in many regions such as cars, roads, guideboards, buildings, the sky, steel bars.
The qualitative results of our visual odometry are shown in Fig. \ref{odometry}.

\subsection{Ablation Study} \label{ablation}
All ablation studies use the same method to train and test DepthNet as Sec. \ref{comparison}. 

\subsubsection{Multi-scale Depth  and  Multiple  Pose Association}
As there are many ways to associate multi-scale depth maps with several refined poses, 
different types of association between poses and depth maps are compared based on 2 PoseNets in the ablation study as shown in Tables III(a). (1) A-depth$\leftrightarrows$A-pose: All depths and all poses are used for losses, and the gradient is calculated for all. (2) A-depth$\leftrightarrows$H-pose, A-depth$\rightarrow$C-pose: All poses are used for losses with all depths, but the gradient is stopped for all depths when associating with coarse poses. 
The results in Tables III(a)  show that "A-depth$\leftrightarrows$H-pose, A-depth$\rightarrow$C-pose" has the best results because this method solves the problem that coarse poses may bring in disturbance to the depth estimation as described in Sec. \ref{multi}.

\subsubsection{Pose Augmentation}
We test the effects of our augmentation methods in the case of 2 PoseNets "A-depth$\leftrightarrows$H-pose, A-depth$\rightarrow$C-pose".
The results in Tables III(b) show that pose augmentation learning brings better performance. Therefore, this pose augmentation method is as the baseline for the following experiments.

\subsubsection{Number of Pose Refinement}

We add more residual PoseNets for more pose refinements and test the results, including 2, 3 and 4 PoseNets, as shown in Tables III(a). These PoseNets all have the same network structure as the PoseNet in Monodepth2~\cite{godard2019digging}. The results show that more pose refinements benefit the results for 1-4 PoseNets. 

\section{Conclusion}\label{con}
The mutual coupling learning based on multi-depth and multi-pose losses is proposed in this paper to refine pose layer by layer. The synthesized middle  view narrows the distance between adjacent frames of images, making residual pose estimation easier, thereby improving the effect of pose estimation. Because of the coupling between depth and pose estimation, depth estimation has also been further improved.
In addition, by synthesizing intermediate view images, 6-DoF pose data augmentation is proposed for this unsupervised learning framework, which enlarges the distribution of pose data. To our knowledge, this paper is the first to realize augmenting the 6-DoF pose in pose learning on 2D images. Finally, our framework achieves state-of-the-art performance on KITTI dataset without any other auxiliary task. 

\ifCLASSOPTIONcaptionsoff
  \newpage
\fi

\bibliographystyle{IEEEtran}  
\bibliography{IEEEabrv,references} 

\begin{thebibliography}{10}
\providecommand{\url}[1]{#1}
\csname url@samestyle\endcsname
\providecommand{\newblock}{\relax}
\providecommand{\bibinfo}[2]{#2}
\providecommand{\BIBentrySTDinterwordspacing}{\spaceskip=0pt\relax}
\providecommand{\BIBentryALTinterwordstretchfactor}{4}
\providecommand{\BIBentryALTinterwordspacing}{\spaceskip=\fontdimen2\font plus
\BIBentryALTinterwordstretchfactor\fontdimen3\font minus
  \fontdimen4\font\relax}
\providecommand{\BIBforeignlanguage}[2]{{%
\expandafter\ifx\csname l@#1\endcsname\relax
\typeout{** WARNING: IEEEtran.bst: No hyphenation pattern has been}%
\typeout{** loaded for the language `#1'. Using the pattern for}%
\typeout{** the default language instead.}%
\else
\language=\csname l@#1\endcsname
\fi
#2}}
\providecommand{\BIBdecl}{\relax}
\BIBdecl

\bibitem{chen2021fixing}
S.~Chen, Z.~Pu, X.~Fan, and B.~Zou, ``Fixing defect of photometric loss for
  self-supervised monocular depth estimation,'' \emph{IEEE Transactions on
  Circuits and Systems for Video Technology}, vol.~32, no.~3, pp. 1328--1338,
  2022.

\bibitem{tian2021depth}
F.~Tian, Y.~Gao, Z.~Fang, Y.~Fang, J.~Gu, H.~Fujita, and J.-N. Hwang, ``Depth
  estimation using a self-supervised network based on cross-layer feature
  fusion and the quadtree constraint,'' \emph{IEEE Transactions on Circuits and
  Systems for Video Technology}, vol.~32, no.~4, pp. 1751--1766, 2022.

\bibitem{wang2021pwclo}
G.~Wang, X.~Wu, Z.~Liu, and H.~Wang, ``Pwclo-net: Deep lidar odometry in 3d
  point clouds using hierarchical embedding mask optimization,'' in
  \emph{Proceedings of the IEEE/CVF Conference on Computer Vision and Pattern
  Recognition}, 2021, pp. 15\,910--15\,919.

\bibitem{eigen2014depth}
D.~Eigen, C.~Puhrsch, and R.~Fergus, ``Depth map prediction from a single image
  using a multi-scale deep network,'' \emph{Advances in Neural Information
  Processing Systems}, vol.~27, pp. 2366--2374, 2014.

\bibitem{song2021monocular}
M.~Song, S.~Lim, and W.~Kim, ``Monocular depth estimation using laplacian
  pyramid-based depth residuals,'' \emph{IEEE transactions on circuits and
  systems for video technology}, vol.~31, no.~11, pp. 4381--4393, 2021.

\bibitem{zhou2017unsupervised}
T.~Zhou, M.~Brown, N.~Snavely, and D.~G. Lowe, ``Unsupervised learning of depth
  and ego-motion from video,'' in \emph{Proceedings of the IEEE Conference on
  Computer Vision and Pattern Recognition}, 2017, pp. 1851--1858.

\bibitem{GANVO}
Y.~{Almalioglu}, M.~R.~U. {Saputra}, P.~P.~B. d.~{Gusmão}, A.~{Markham}, and
  N.~{Trigoni}, ``Ganvo: Unsupervised deep monocular visual odometry and depth
  estimation with generative adversarial networks,'' in \emph{2019
  International Conference on Robotics and Automation (ICRA)}, 2019, pp.
  5474--5480.

\bibitem{8793622}
G.~Wang, H.~Wang, Y.~Liu, and W.~Chen, ``Unsupervised learning of monocular
  depth and ego-motion using multiple masks,'' in \emph{2019 International
  Conference on Robotics and Automation (ICRA)}, 2019, pp. 4724--4730.

\bibitem{bian2019depth}
J.-W. Bian, Z.~Li, N.~Wang, H.~Zhan, C.~Shen, M.-M. Cheng, and I.~Reid,
  ``Unsupervised scale-consistent depth and ego-motion learning from monocular
  video,'' in \emph{Thirty-third Conference on Neural Information Processing
  Systems (NeurIPS)}, 2019.

\bibitem{godard2019digging}
C.~Godard, O.~Mac~Aodha, M.~Firman, and G.~J. Brostow, ``Digging into
  self-supervised monocular depth estimation,'' in \emph{Proceedings of the
  IEEE/CVF International Conference on Computer Vision}, 2019, pp. 3828--3838.

\bibitem{Gordon_2019_ICCV}
A.~Gordon, H.~Li, R.~Jonschkowski, and A.~Angelova, ``Depth from videos in the
  wild: Unsupervised monocular depth learning from unknown cameras,'' in
  \emph{Proceedings of the IEEE/CVF International Conference on Computer Vision
  (ICCV)}, October 2019.

\bibitem{yang2018unsupervised}
Z.~Yang, P.~Wang, W.~Xu, L.~Zhao, and R.~Nevatia, ``Unsupervised learning of
  geometry from videos with edge-aware depth-normal consistency,'' in
  \emph{Proceedings of the AAAI Conference on Artificial Intelligence},
  vol.~32, no.~1, 2018.

\bibitem{Yin_2018_CVPR}
Z.~Yin and J.~Shi, ``Geonet: Unsupervised learning of dense depth, optical flow
  and camera pose,'' in \emph{Proceedings of the IEEE Conference on Computer
  Vision and Pattern Recognition (CVPR)}, June 2018.

\bibitem{Zou_2018_ECCV}
Y.~Zou, Z.~Luo, and J.-B. Huang, ``Df-net: Unsupervised joint learning of depth
  and flow using cross-task consistency,'' in \emph{Proceedings of the European
  Conference on Computer Vision (ECCV)}, September 2018.

\bibitem{Ranjan_2019_CVPR}
A.~Ranjan, V.~Jampani, L.~Balles, K.~Kim, D.~Sun, J.~Wulff, and M.~J. Black,
  ``Competitive collaboration: Joint unsupervised learning of depth, camera
  motion, optical flow and motion segmentation,'' in \emph{Proceedings of the
  IEEE/CVF Conference on Computer Vision and Pattern Recognition (CVPR)}, June
  2019.

\bibitem{wang2020unsupervised}
G.~Wang, C.~Zhang, H.~Wang, J.~Wang, Y.~Wang, and X.~Wang, ``Unsupervised
  learning of depth, optical flow and pose with occlusion from 3d geometry,''
  \emph{IEEE Transactions on Intelligent Transportation Systems}, 2020.

\bibitem{9156629}
W.~{Zhao}, S.~{Liu}, Y.~{Shu}, and Y.~J. {Liu}, ``Towards better
  generalization: Joint depth-pose learning without posenet,'' in \emph{2020
  IEEE/CVF Conference on Computer Vision and Pattern Recognition (CVPR)}, 2020,
  pp. 9148--9158.

\bibitem{doi:10.1177/0278364913491297}
\BIBentryALTinterwordspacing
A.~Geiger, P.~Lenz, C.~Stiller, and R.~Urtasun, ``Vision meets robotics: The
  kitti dataset,'' \emph{The International Journal of Robotics Research},
  vol.~32, no.~11, pp. 1231--1237, 2013. [Online]. Available:
  \url{https://doi.org/10.1177/0278364913491297}
\BIBentrySTDinterwordspacing

\bibitem{wang2020tartanair}
W.~Wang, D.~Zhu, X.~Wang, Y.~Hu, Y.~Qiu, C.~Wang, Y.~Hu, A.~Kapoor, and
  S.~Scherer, ``Tartanair: A dataset to push the limits of visual slam,'' in
  \emph{2020 IEEE/RSJ International Conference on Intelligent Robots and
  Systems (IROS)}.\hskip 1em plus 0.5em minus 0.4em\relax IEEE, 2020, pp.
  4909--4916.

\bibitem{tartanvo2020corl}
W.~Wang, Y.~Hu, and S.~Scherer, ``Tartanvo: A generalizable learning-based
  vo,'' in \emph{Conference on Robot Learning (CoRL)}, 2020.

\bibitem{mur2017orb}
R.~Mur-Artal and J.~D. Tard{\'o}s, ``Orb-slam2: An open-source slam system for
  monocular, stereo, and rgb-d cameras,'' \emph{IEEE Transactions on Robotics},
  vol.~33, no.~5, pp. 1255--1262, 2017.

\bibitem{liu2015learning}
F.~Liu, C.~Shen, G.~Lin, and I.~Reid, ``Learning depth from single monocular
  images using deep convolutional neural fields,'' \emph{IEEE Transactions on
  Pattern Analysis and Machine Intelligence}, vol.~38, no.~10, pp. 2024--2039,
  2015.

\bibitem{Kuznietsov_2017_CVPR}
Y.~Kuznietsov, J.~Stuckler, and B.~Leibe, ``Semi-supervised deep learning for
  monocular depth map prediction,'' in \emph{Proceedings of the IEEE Conference
  on Computer Vision and Pattern Recognition (CVPR)}, July 2017.

\bibitem{Guo_2018_ECCV}
X.~Guo, H.~Li, S.~Yi, J.~Ren, and X.~Wang, ``Learning monocular depth by
  distilling cross-domain stereo networks,'' in \emph{Proceedings of the
  European Conference on Computer Vision (ECCV)}, September 2018.

\bibitem{garg2016unsupervised}
R.~Garg, V.~K. Bg, G.~Carneiro, and I.~Reid, ``Unsupervised cnn for single view
  depth estimation: Geometry to the rescue,'' in \emph{European Conference on
  Computer Vision}.\hskip 1em plus 0.5em minus 0.4em\relax Springer, 2016, pp.
  740--756.

\bibitem{godard2017unsupervised}
C.~Godard, O.~Mac~Aodha, and G.~J. Brostow, ``Unsupervised monocular depth
  estimation with left-right consistency,'' in \emph{Proceedings of the IEEE
  Conference on Computer Vision and Pattern Recognition}, 2017, pp. 270--279.

\bibitem{zhan2018unsupervised}
H.~Zhan, R.~Garg, C.~S. Weerasekera, K.~Li, H.~Agarwal, and I.~Reid,
  ``Unsupervised learning of monocular depth estimation and visual odometry
  with deep feature reconstruction,'' in \emph{Proceedings of the IEEE
  Conference on Computer Vision and Pattern Recognition}, 2018, pp. 340--349.

\bibitem{8747446}
T.~{Feng} and D.~{Gu}, ``Sganvo: Unsupervised deep visual odometry and depth
  estimation with stacked generative adversarial networks,'' \emph{IEEE
  Robotics and Automation Letters}, vol.~4, no.~4, pp. 4431--4437, 2019.

\bibitem{wong2019bilateral}
A.~Wong and S.~Soatto, ``Bilateral cyclic constraint and adaptive
  regularization for unsupervised monocular depth prediction,'' in
  \emph{Proceedings of the IEEE/CVF Conference on Computer Vision and Pattern
  Recognition}, 2019, pp. 5644--5653.

\bibitem{luo2019every}
C.~Luo, Z.~Yang, P.~Wang, Y.~Wang, W.~Xu, R.~Nevatia, and A.~Yuille, ``Every
  pixel counts++: Joint learning of geometry and motion with 3d holistic
  understanding,'' \emph{IEEE Transactions on Pattern Analysis and Machine
  Intelligence}, vol.~42, no.~10, pp. 2624--2641, 2019.

\bibitem{Yang_2020_CVPR}
N.~Yang, L.~v. Stumberg, R.~Wang, and D.~Cremers, ``D3vo: Deep depth, deep pose
  and deep uncertainty for monocular visual odometry,'' in \emph{Proceedings of
  the IEEE/CVF Conference on Computer Vision and Pattern Recognition (CVPR)},
  June 2020.

\bibitem{yang2018lego}
Z.~Yang, P.~Wang, Y.~Wang, W.~Xu, and R.~Nevatia, ``Lego: Learning edge with
  geometry all at once by watching videos,'' in \emph{Proceedings of the IEEE
  Conference on Computer Vision and Pattern Recognition}, 2018, pp. 225--234.

\bibitem{shen2019beyond}
T.~Shen, Z.~Luo, L.~Zhou, H.~Deng, R.~Zhang, T.~Fang, and L.~Quan, ``Beyond
  photometric loss for self-supervised ego-motion estimation,'' in \emph{2019
  International Conference on Robotics and Automation (ICRA)}.\hskip 1em plus
  0.5em minus 0.4em\relax IEEE, 2019, pp. 6359--6365.

\bibitem{mahjourian2018unsupervised}
R.~Mahjourian, M.~Wicke, and A.~Angelova, ``Unsupervised learning of depth and
  ego-motion from monocular video using 3d geometric constraints,'' in
  \emph{Proceedings of the IEEE Conference on Computer Vision and Pattern
  Recognition}, 2018, pp. 5667--5675.

\bibitem{casser2019depth}
V.~Casser, S.~Pirk, R.~Mahjourian, and A.~Angelova, ``Depth prediction without
  the sensors: Leveraging structure for unsupervised learning from monocular
  videos,'' in \emph{Proceedings of the AAAI conference on artificial
  intelligence}, vol.~33, no.~01, 2019, pp. 8001--8008.

\bibitem{9345430}
Y.~{Liu}, H.~{Wang}, J.~{Wang}, and X.~{Wang}, ``Unsupervised monocular visual
  odometry based on confidence evaluation,'' \emph{IEEE Transactions on
  Intelligent Transportation Systems}, pp. 1--10, 2021.

\bibitem{wang2018learning}
C.~Wang, J.~M. Buenaposada, R.~Zhu, and S.~Lucey, ``Learning depth from
  monocular videos using direct methods,'' in \emph{Proceedings of the IEEE
  Conference on Computer Vision and Pattern Recognition}, 2018, pp. 2022--2030.

\bibitem{zhao2019geometry}
S.~Zhao, H.~Fu, M.~Gong, and D.~Tao, ``Geometry-aware symmetric domain
  adaptation for monocular depth estimation,'' in \emph{Proceedings of the
  IEEE/CVF Conference on Computer Vision and Pattern Recognition}, 2019, pp.
  9788--9798.

\bibitem{li2020unsupervised}
H.~Li, A.~Gordon, H.~Zhao, V.~Casser, and A.~Angelova, ``Unsupervised monocular
  depth learning in dynamic scenes,'' \emph{Conference on Robot Learning
  (CoRL)}, 2020.

\bibitem{zhang2021dpsnet}
J.~Zhang, Q.~Su, B.~Tang, C.~Wang, and Y.~Li, ``Dpsnet: Multitask learning
  using geometry reasoning for scene depth and semantics,'' \emph{IEEE
  Transactions on Neural Networks and Learning Systems}, 2021.

\bibitem{wang2004image}
Z.~Wang, A.~C. Bovik, H.~R. Sheikh, and E.~P. Simoncelli, ``Image quality
  assessment: from error visibility to structural similarity,'' \emph{IEEE
  Transactions on Image Processing}, vol.~13, no.~4, pp. 600--612, 2004.

\bibitem{tosi2020distilled}
F.~Tosi, F.~Aleotti, P.~Z. Ramirez, M.~Poggi, S.~Salti, L.~D. Stefano, and
  S.~Mattoccia, ``Distilled semantics for comprehensive scene understanding
  from videos,'' in \emph{Proceedings of the IEEE/CVF Conference on Computer
  Vision and Pattern Recognition}, 2020, pp. 4654--4665.

\bibitem{jia2021self}
S.~Jia, X.~Pei, X.~Jing, and D.~Yao, ``Self-supervised 3d reconstruction and
  ego-motion estimation via on-board monocular video,'' \emph{IEEE Transactions
  on Intelligent Transportation Systems}, 2021.

\bibitem{10.1007/978-3-030-58452-8_42}
J.~Watson, O.~M. Aodha, D.~Turmukhambetov, G.~J. Brostow, and M.~Firman,
  ``Learning stereo from single images,'' in \emph{Computer Vision -- ECCV
  2020}, A.~Vedaldi, H.~Bischof, T.~Brox, and J.-M. Frahm, Eds.\hskip 1em plus
  0.5em minus 0.4em\relax Cham: Springer International Publishing, 2020, pp.
  722--740.

\bibitem{Aleotti_2021_CVPR}
F.~Aleotti, M.~Poggi, and S.~Mattoccia, ``Learning optical flow from still
  images,'' in \emph{Proceedings of the IEEE/CVF Conference on Computer Vision
  and Pattern Recognition (CVPR)}, June 2021, pp. 15\,201--15\,211.

\bibitem{telea2004image}
A.~Telea, ``An image inpainting technique based on the fast marching method,''
  \emph{Journal of Graphics Tools}, vol.~9, no.~1, pp. 23--34, 2004.

\bibitem{kingma2014adam}
D.~P. Kingma and J.~Ba, ``Adam: A method for stochastic optimization,''
  \emph{arXiv preprint arXiv:1412.6980}, 2014.

\end{thebibliography}

	\vspace{-10mm}
\begin{IEEEbiography}[{\includegraphics[width=1in,height=1.25in,clip,keepaspectratio]{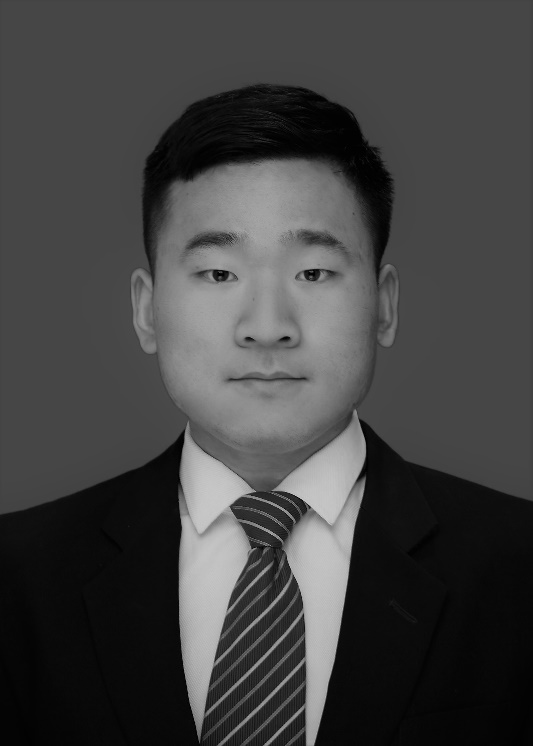}}]{Guangming Wang}
received the B.S. degree from Department of Automation from Central South University, Changsha, China, in 2018. He is currently pursuing the Ph.D. degree in Control Science and Engineering with Shanghai Jiao Tong University. His current research interests include SLAM and computer vision, in particular, 2D optical flow estimation and 3D scene flow estimation.
\end{IEEEbiography}
	\vspace{-10mm}
\begin{IEEEbiography}[{\includegraphics[width=0.9in,height=1.3in,clip]{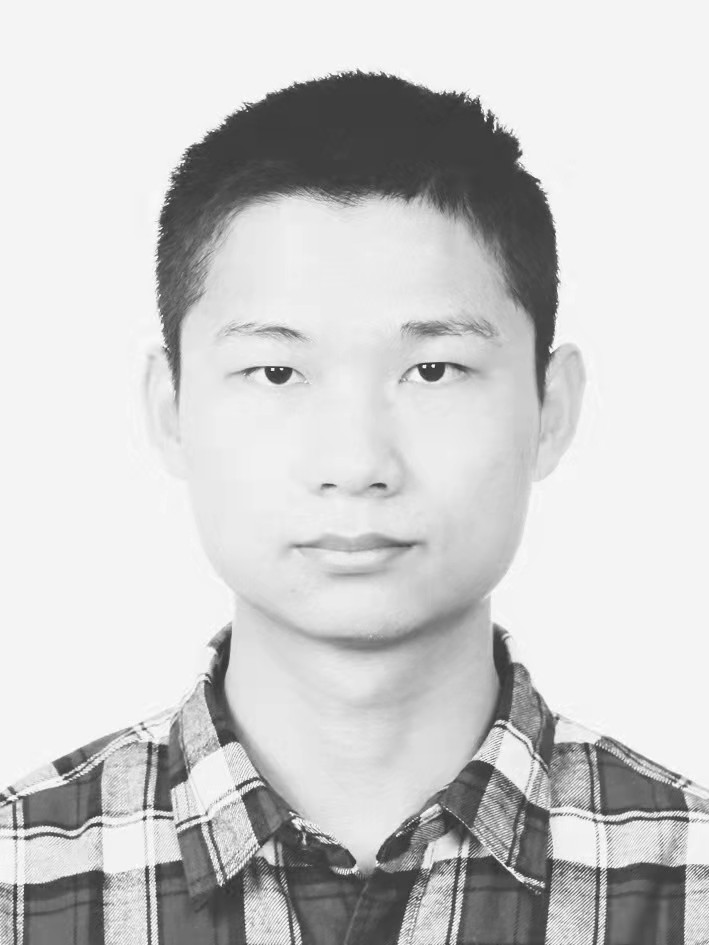}}]{Jiquan Zhong}
is currently pursuing the B.S. degree in Department of Automation, Shanghai Jiao Tong University. His latest research interests include SLAM and computer vision. 
\end{IEEEbiography}
\begin{IEEEbiography}[{\includegraphics[width=0.9in,height=1.3in,clip]{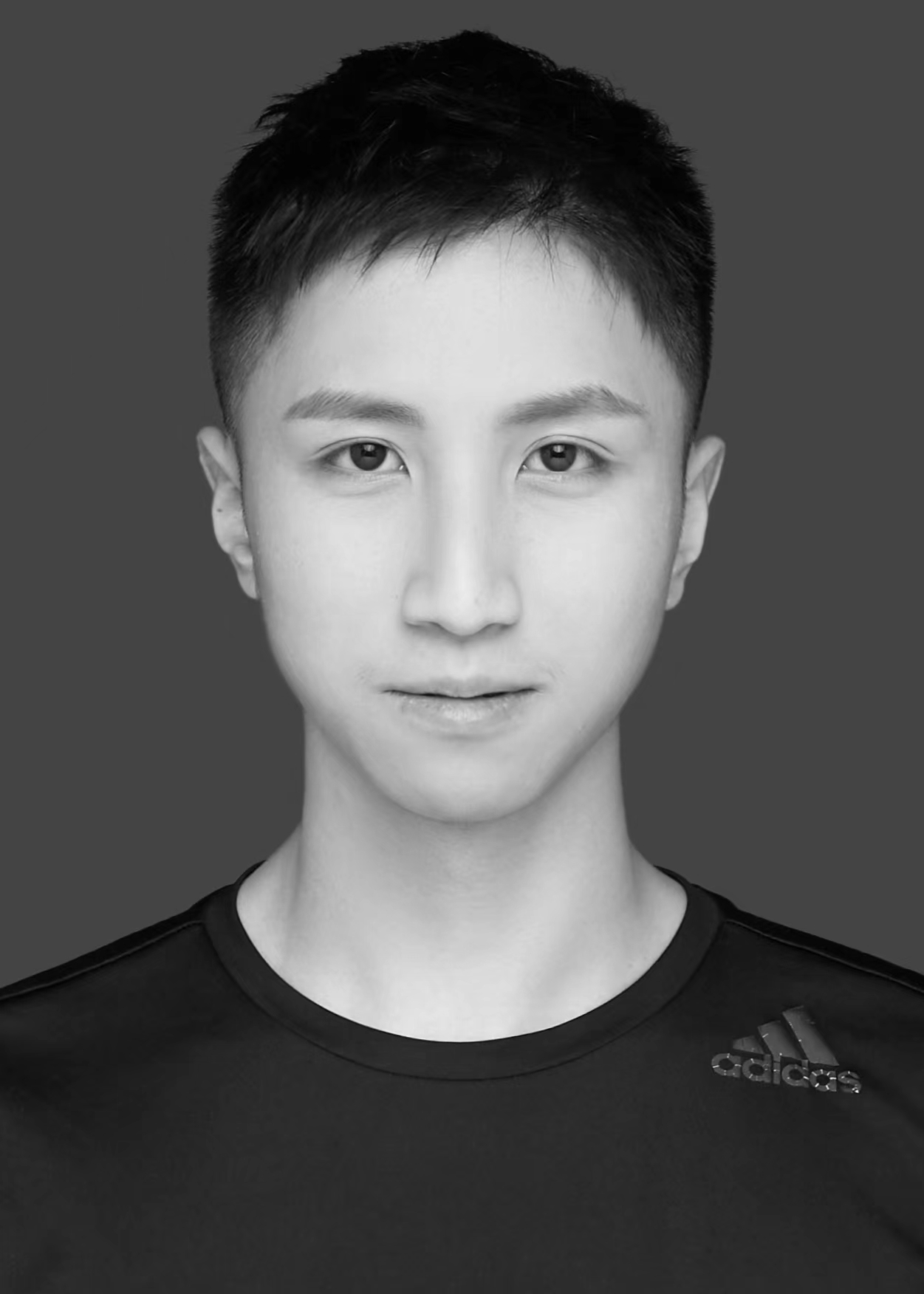}}]{Shijie Zhao}
is currently pursuing the B.S. degree in Department of Engineering Mechanics, Shanghai Jiao Tong University. His latest research interests include SLAM and computer vision. 
\end{IEEEbiography}
\begin{IEEEbiography}[{\includegraphics[width=0.9in,height=1.3in,clip]{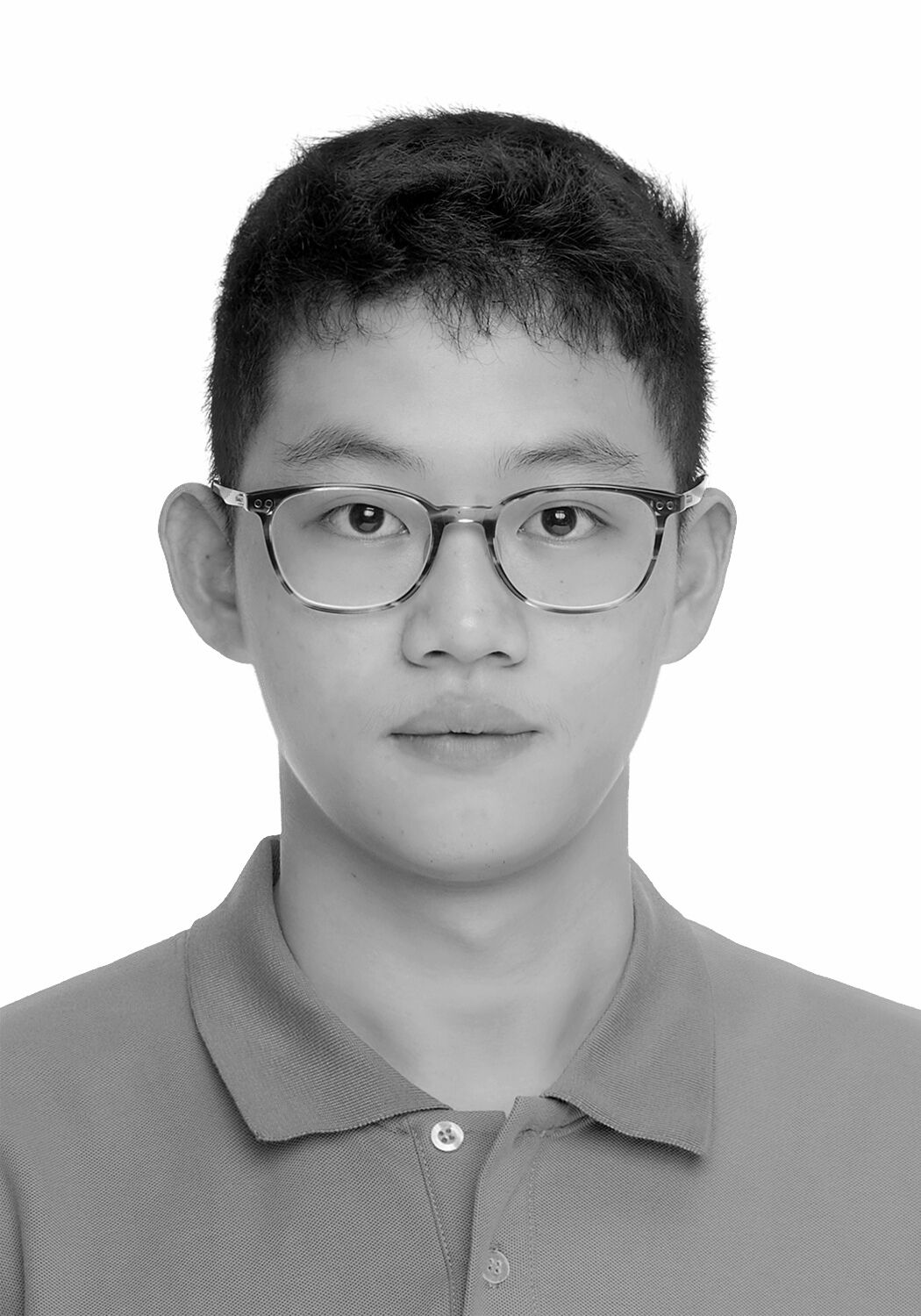}}]{Wenhua}
is currently pursuing the B.S. degree in Department of Automation, Shanghai Jiao Tong University. His latest research interests include SLAM and computer vision. 
\end{IEEEbiography}
\begin{IEEEbiography}[{\includegraphics[width=1in,height=1.25in,clip,keepaspectratio]{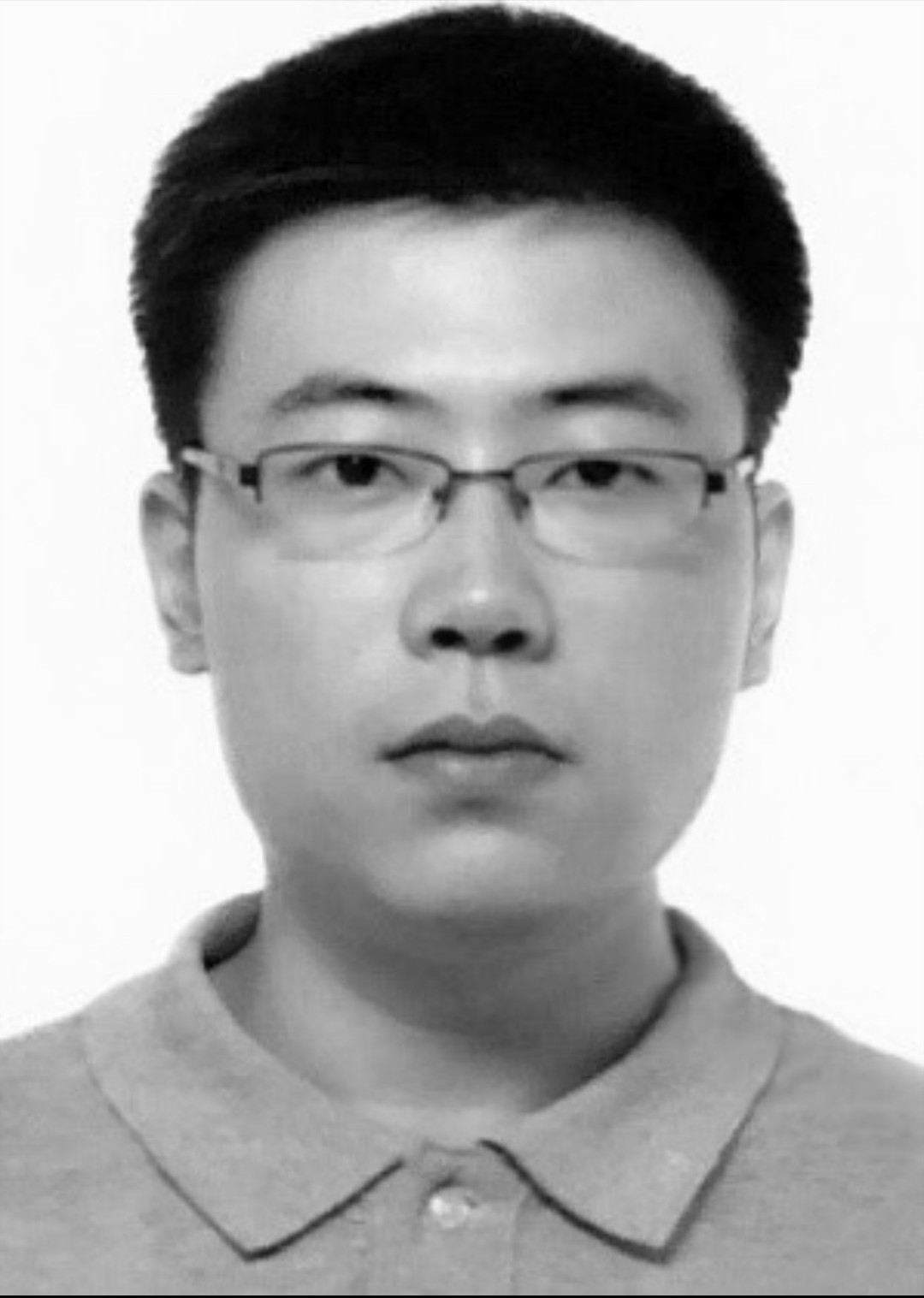}}]{Zhe Liu} received his B.S. degree in Automation from Tianjin University, Tianjin, China, in 2010, and Ph.D. degree in Control Technology and Control Engineering from Shanghai Jiao Tong University, Shanghai, China, in 2016. From 2017 to 2020, he was a Post-Doctoral Fellow with the Department of Mechanical and Automation Engineering, The Chinese University of Hong Kong, Hong Kong. He is currently a Research Associate with the Department of Computer Science and Technology, University of Cambridge. His research interests include autonomous mobile robot, multirobot cooperation and autonomous driving system. 
\end{IEEEbiography}
\begin{IEEEbiography}[{\includegraphics[width=1in,height=1.25in,clip,keepaspectratio]{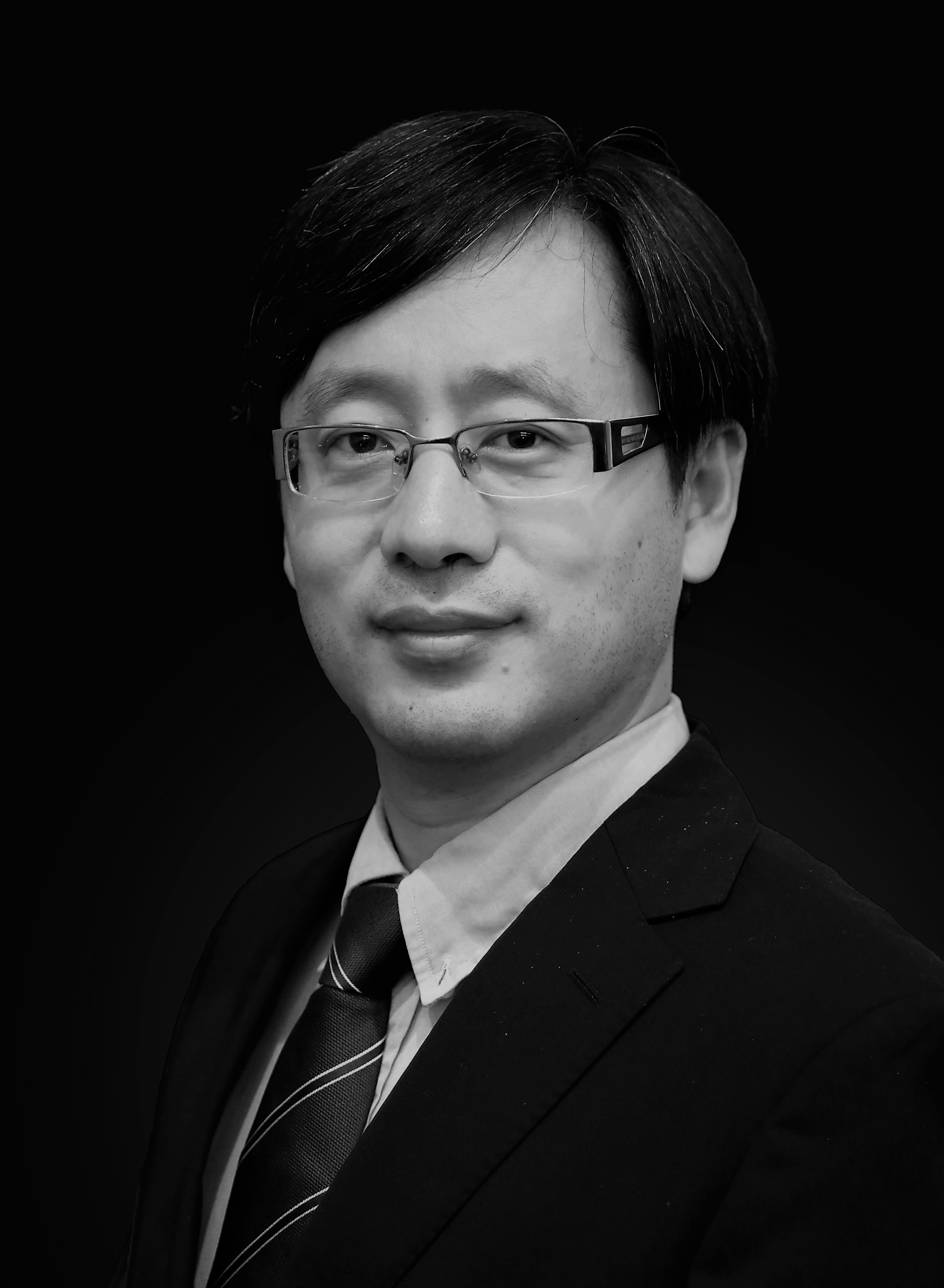}}]{Hesheng Wang}
received the B.Eng. degree in electrical engineering from the Harbin Institute of Technology, Harbin, China, in 2002, and the M.Phil. and Ph.D. degrees in automation and computer-aided engineering from The Chinese University of Hong Kong, Hong Kong, in 2004 and 2007, respectively. He is currently a Professor with the Department of Automation, Shanghai Jiao Tong University, Shanghai, China. His current research interests include visual servoing, service robot, computer vision, and autonomous driving. 
Dr. Wang is an Associate Editor of IEEE Transactions on Automation Science and Engineering, IEEE Robotics and Automation Letters, Assembly Automation and the International Journal of Humanoid Robotics, a Technical Editor of the IEEE/ASME Transactions on Mechatronics, an Editor of Conference Editorial Board (CEB) of IEEE Robotics and Automation Society. He served as an Associate Editor of the IEEE Transactions on Robotics from 2015 to 2019. He was the General Chair of the IEEE RCAR 2016, and the Program Chair of the IEEE ROBIO 2014 and IEEE/ASME AIM 2019. He is the General Chair of IEEE ROBIO 2022.
\end{IEEEbiography}

\end{document}